\documentclass[lettersize,journal]{IEEEtran}
\usepackage{amsmath,amsfonts}
\usepackage{algorithmic}
\usepackage{algorithm}
\usepackage{array}
\usepackage[caption=false,font=normalsize,labelfont=sf,textfont=sf]{subfig}
\usepackage{textcomp}
\usepackage{xcolor}
\usepackage{stfloats}
\usepackage{url}
\usepackage{verbatim}
\usepackage{graphicx}
\usepackage{cite}
\usepackage{graphicx}
\usepackage{float} 
\usepackage{hyperref}
\usepackage{cleveref}
\usepackage[utf8]{inputenc}
\usepackage{orcidlink}
\begin{document}

\title{SR-SLAM: Scene Reliability-Based RGB-D SLAM in Diverse Environments}

\author{Haolan Zhang, \IEEEmembership{Student member,~IEEE}, Chenghao Li~\orcidlink{0009-0003-0404-3825}, \IEEEmembership{Gradute Student member,~IEEE}, Thanh Nguyen Canh~\orcidlink{0000-0001-6332-1002},~\IEEEmembership{Gradute Student member, IEEE},  Lijun Wang, \IEEEmembership{Gradute Student member,~IEEE}, Nak Young Chong~\orcidlink{0000-0001-5736-0769}, \IEEEmembership{Senior member, IEEE}
\thanks{}
\thanks{The authors are with the School of Information Science, Japan Advanced Institute of Science and Technology, Nomi 923-1211, Japan (email: \{haolan.z; chenghao.li; thanhnc; lijun.wang; nakyoung\}@jaist.ac.jp).}
\thanks{}}


\markboth{}%
{Shell \MakeLowercase{\textit{et al.}}: A Sample Article Using IEEEtran.cls for IEEE Journals}


\maketitle

\begin{abstract}
Visual simultaneous localization and mapping (SLAM) plays a critical role in autonomous robotic systems, especially where accurate and reliable measurements are essential for navigation and sensing. In feature-based SLAM, the quantity and quality of extracted features significantly influence system performance. Due to the variations in feature quantity and quality across diverse environments, current approaches face two major challenges:  (1) limited adaptability in dynamic feature culling and pose estimation, and (2) insufficient environmental awareness in assessment and optimization strategies. To address these issues, we propose SRR-SLAM, a scene-reliability based framework that enhances feature-based SLAM through environment-aware processing. Our method introduces a unified scene reliability assessment mechanism that incorporates multiple metrics and historical observations to guide system behavior. Based on this assessment, we develop: (i) adaptive dynamic region selection with flexible geometric constraints, (ii) depth-assisted self-adjusting clustering for efficient dynamic feature removal in high-dimensional settings, and (iii) reliability-aware pose refinement that dynamically integrates direct methods when features are insufficient. Furthermore, we propose (iv) reliability-based keyframe selection and a weighted optimization scheme to reduce computational overhead while improving estimation accuracy. Extensive experiments on public datasets and real-world scenarios show that SRR-SLAM outperforms state-of-the-art dynamic SLAM methods, achieving up to 90\% improvement in accuracy and robustness across diverse environments. These improvements directly contribute to enhanced measurement precision and reliability in autonomous robotic sensing systems.
\end{abstract}


\section{Introduction}
\IEEEPARstart{S}{LAM} is a core component of autonomous systems. Among various modalities, visual SLAM~\cite{survey1} has received increasing attention due to the cost-effectiveness and rich environmental information provided by cameras. Visual SLAM techniques are typically categorized into direct methods, which exploit pixel intensity information, and feature-based methods, which rely on the detection and matching of distinctive feature points. Feature-based methods are widely adopted due to their computational efficiency and strong performance in texture-rich environments. The effectiveness of feature-based SLAM depends heavily on two key factors: the quantity and quality of extracted features, which directly influence the accuracy and reliability of pose estimation. However, both aspects face significant challenges in diverse environments - ranging from completely static scenarios to dynamic scenarios of varying complexity. In these diverse environments, quantity and quality of available features are compromised by multiple factors: inherent scene texture limitations, dynamic interference, and occlusions from both static and dynamic objects. Such degradation results in feature insufficiency that undermines pose estimation reliability and poor feature quality that introduces measurement uncertainty, both of which can severely compromise overall system performance.

From the feature quantity perspective, insufficient features pose critical challenges for both traditional and dynamic SLAM systems. Traditional feature-based SLAM methods ORB-SLAM family~\cite{ORB-SLAM,ORB-SLAM2,ORB-SLAM3} encounter challenges with limited features due to occlusion, low-texture environments, which can degrade pose estimation accuracy or cause tracking failures. In dynamic SLAM methods~\cite{survey2}, these quantity challenges are severely compounded by moving objects that bring dynamic interference and reduce available static features. While researchers have developed potential dynamic region computation and dynamic feature removal strategies to identify and eliminate dynamic interference, reduced available features still create a cascade of problems ranging from unreliable geometric computations to insufficient pose estimation. First, regarding unreliable geometric computations, methods such as~\cite{DS-SLAM, DynaSLAM, BlitzSLAM} apply geometric constraints in fixed ways during dynamic region computation. These strategies become fundamentally unreliable when the number of reliable features drops, often leading to incorrect region identification and residual dynamic interference. Second, for insufficient pose estimation, some methods focus on preserving more usable features using proportion-based thresholds~\cite{DRG-SLAM, DPL-SLAM} or clustering algorithms~\cite{RSO-SLAM, SSF-SLAM, HMC-SLAM, forest 3-D, DBSCAN-Based Lidar, DGMVINS}. Others attempt to complement feature-based estimation by integrating direct methods~\cite{semi-dense, Tight integration, UniVIO}. While these efforts partially address the issue, they still face significant limitations: a lack of robustness against non-rigid object deformations, poor handling of high-dimensional data, and reliance on fixed parameter settings without environmental adaptivity in algorithm applications. Moreover, fusion between direct and feature-based methods remains static, lacking dynamic adjustment based on real-time scene conditions.

Beyond the feature quantity limitations discussed above, recent research has focused on environment assessment based adaptive processing strategies from the feature quality. However, existing approaches suffer from three critical limitations. First, quality assessment methods~\cite{switching-coupled} lack temporal continuity, performing independent frame-wise evaluations without considering historical consistency and environmental evolution. Second, optimization processes incur excessive computational complexity through fine-grained feature-level weighting~\cite{HMC-SLAM, QualiSLAM} that necessitates individual treatment of each feature point. Third, frame-level processing~\cite{PointSlot} employs unrealistic binary classification that ignores actual environmental conditions, failing to account for varying feature quality across different scenarios. These limitations result in suboptimal performance in diverse environments where feature quality varies significantly over time.

To address the aforementioned limitations in both feature quantity and feature quality, this paper proposes SRR-SLAM, a scene-reliability based RGB-D SLAM framework that achieves unified assessment of current frames through multiple metrics (object detection confidence and spatial distribution, feature quality, depth quality) and flexible historical references, resolving the lack of continuity and adaptation in existing assessment processes. 

Based on this reliability assessment, our framework guides adaptive strategies across five key components: dynamic region selection, dynamic feature removal, pose refinement, new keyframe selection, and weighted optimization. For dynamic region selection and removal, we dynamically determine the level of scene reliability to flexibly select appropriate geometric constraints, thereby addressing the fixed application of geometric constraints in prior methods. Subsequently, dynamic removal is performed using our depth-assisted adaptive DBSCAN (Density-Based Spatial Clustering of Applications with Noise): initially, depth information is used to pre-cluster features to handle high-dimensional data, then the adaptive DBSCAN adjusts its parameters according to the assessment results to complete the removal process. This approach effectively addresses challenges related to non-rigid objects and the need for self-adjustment in removal algorithms. For pose refinement, direct methods are integrated to complement feature-based pose estimation exclusively in low-reliability scenarios. The fusion process dynamically adjusts the weighting of the pose estimates from both methods based on the current environmental reliability to obtain the fused pose, effectively addressing poor estimation caused by insufficient features and overcoming the inflexibility of existing direct-feature fusion methods across diverse environments. Finally, for keyframe selection and optimization, we implement a new keyframe selection strategy alongside frame-level weighted optimization based on the reliability assessment to ensure the reliability of selected keyframes and focus optimization on high-reliability frames. This resolves the computational complexity of feature-level operations and the unrealistic frame classification strategies that ignore actual frame conditions.

This article addresses fundamental challenges in visual robotic sensing systems. Our unified reliability assessment framework advances the state-of-the-art in measurement quality evaluation and the reliability based strategies enhance measurement accuracy and robustness of vision sensors in diverse environmental. These contributions are particularly valuable for precision robotics applications where sensor measurement reliability directly impacts system performance and safety.

The main contributions of this article include the following.
\begin{enumerate} 
    \item 
    A multi-faceted and temporally continuous scene reliability assessment mechanism to quantitatively evaluate the environmental reliability of each frame for feature-based methods.
    \item An adaptive dynamic culling strategy based on scene reliability that adaptively applies geometric constraints to select dynamic regions and employs depth-assisted adaptive DBSCAN with self adjustment.
    \item A reliability-aware pose refinement, new keyframe selection, and optimization mechanism. When scene reliability is low, direct methods are integrated for pose refinement, along with reliability-based keyframe selection and optimization.
    \item Extensive comparative experiments on public datasets (TUM~\cite{TUM} and BONN~\cite{BONN}) demonstrating significant improvements over state-of-the-art dynamic SLAM methods and trajectory error reduction compared to our baseline ORB-SLAM3 in real scenarios.
\end{enumerate}

The remainder of the paper is organized as follows. Section \uppercase\expandafter{\romannumeral2} presents the related work associated with feature-based SLAM and scene assessment approaches. Section \uppercase\expandafter{\romannumeral3} shows the details related to the methods of SRR-SLAM. Section \uppercase\expandafter{\romannumeral4} provides experimental evaluation and analysis. The conclusion and future works are presented in Section \uppercase\expandafter{\romannumeral5}.

\section{RELATED WORKS}
The method proposed in this article aims to enhance the accuracy and robustness of feature-based approaches in diverse environments by introducing a unified assessment framework with temporal continuity and environmental adaptability. This framework guides flexible dynamic processing, pose refinement when available features are insufficient, and enables assessment-based keyframe management and optimization for downstream processing. To better position our work within the existing literature and highlight current research gaps, related methods are categorized into three core areas aligned with our main contributions: dynamic processing, pose refinement with direct methods, and assessment-based methods.

\subsection{Dynamic Processing}
The process of dynamic interference removal consists of two main steps: dynamic region computation and dynamic feature removal. For dynamic region computation, 
pure geometric approaches~\cite{point weighting, point correlations, Remote Sens} have evolved to integrate deep learning detection with geometric constraints. DS-SLAM deploys SegNet~\cite{SegNet} in separate threads for accelerated object detection while applying optical flow consistency checks to all feature points. DynaSLAM combines Mask R-CNN~\cite{Mask R-CNN} detection results with multi-view geometry~\cite{Multiple View Geometry} for dynamic feature selection and removal. YOLO-SLAM~\cite{YOLO-SLAM} utilizes YOLO for dynamic object detection, followed by depth-RANSAC~\cite{RANSAC} combined with depth information for dynamic feature removal within object regions. Blitz-SLAM applies BlitzNet~\cite{BlitzNet} to obtain semantic masks, then uses epipolar geometry to identify dynamic features in potential dynamic regions while refining masks using depth information in separate threads to generate improved local point clouds. For dynamic feature removal, numerous methods have emerged for the flexible processing of features after the dynamic region selection. Threshold-based methods such as DRG-SLAM~\cite{DRG-SLAM} and DPL-SLAM~\cite{DPL-SLAM} determine region-level removal strategies based on dynamic feature proportions within detected areas. In addition, clustering methods have gradually been introduced into the process of dynamic point removal. For example, RSO-SLAM~\cite{RSO-SLAM} employs k-means~\cite{k-means} clustering to refine optical flow results, SSF-SLAM~\cite{SSF-SLAM} uses depth-based convex hull with motion consistency evaluation, and HMC-SLAM~\cite{HMC-SLAM} integrates Hierarchical Multidimensional Clustering (HMC) with geometric constraints. DBSCAN has gained popularity in dynamic processing applications~\cite{forest 3-D,DBSCAN-Based Lidar,DGMVINS} due to its density-based clustering capabilities and automatic cluster determination, enabling effective separation of geometrically consistent and inconsistent features without predefined cluster shapes.

While these approaches have achieved considerable success in dynamic scenarios, they suffer from fixed applications with geometric constraints and lack environmental adaptivity. 
The integration of deep learning and geometric verification~\cite{DS-SLAM, DynaSLAM, YOLO-SLAM, BlitzSLAM} mechanically applies identical geometric constraints for dynamic region selection regardless of environmental variations. When available static features fall below minimum thresholds required for reliable fundamental matrix or essential matrix computation, these geometric constraints fail completely, leading to system breakdown in challenging scenarios with poor texture or dynamic occlusions. Some approaches~\cite{DRG-SLAM, DPL-SLAM, RSO-SLAM, SSF-SLAM, HMC-SLAM} rely on fixed parameters and thresholds that cannot self-adjust to varying scene characteristics. DBSCAN-based methods~\cite{forest 3-D,DBSCAN-Based Lidar,DGMVINS} still require manual parameter tuning and suffer from performance degradation with high-dimensional data, creating accuracy trade-offs in complex scenes with non-uniform density.
\subsection{Pose Refinement with Direct Methods}
To achieve measurement accuracy in low-feature environments, several methods have explored utilizing direct methods that do not require feature correspondences to compensate for the limitations of current feature-based estimation. A real-time semi-dense stereo SLAM system~\cite{semi-dense} with feature-based odometry prior was proposed to combine the advantages of both approaches. Other works~\cite{Tight integration} have integrated feature point matching with semi-dense direct image alignment to achieve robust and fast tracking performance. UniVIO~\cite{UniVIO} merges direct methods and feature point techniques in situations with fluctuating illumination conditions, thereby augmenting system adaptability by concurrently contributing to visual residuals through both photometric and geometric constraints.

However, existing direct method integration approaches typically employ fixed combination strategies without intelligent fusing mechanisms based on environmental conditions. This often results in the direct method becoming a source of interference when feature-based methods are sufficient. Conversely, when the performance of feature-based methods deteriorates, although direct methods are introduced as a supplement, there is a lack of an adaptive mechanism to determine the extent of compensation required based on the current scene. This highlights the absence of a dynamic balance between the two methods.
\subsection{Assessment-based methods}
Assessment-based approaches that evaluate various environmental factors in the current environment to adopt different application strategies have become the primary methodology. Recent approaches have explored adaptive strategies that integrate assessment and optimization at feature and frame levels. At the feature level, 
feature ranking systems were proposed based on observation scores for pose estimation contribution~\cite{Good features} and adaptive feature extraction thresholds were employed based on illumination conditions~\cite{Illumination-Adaptive}. The switching-coupled approach~\cite{switching-coupled} evaluates objects based on uncertainty, observation quality, and prior information with every frame independently, then determines whether to apply tightly-coupled or loosely-coupled methods for pose estimation. QualiSLAM~\cite{QualiSLAM} employs six evaluation metrics, such as illumination, contrast, and texture to assess each scene block, then applies the CRITIC weighting method to calculate block weights according to the scene evaluation. It uses these weights during pose optimization and Bundle Adjustment with features to mitigate challenging scenarios. HMC-SLAM~\cite{HMC-SLAM} detects residual dynamic features through depth residual analysis, then applies feature weighting based on average depth differences between detected region features and maximum dynamic clusters to reduce the influence of potentially missed dynamic features in joint optimization. At the frame level, PointSLOT~\cite{PointSlot} introduces a keyframe management strategy that separates object and background keyframes based on the presence of dynamic objects. This separation enables independent object pose estimation relative to the camera, thereby eliminating the interference of camera pose estimation on object tracking.

Despite achieving certain levels of flexible processing through assessment-based strategies, these approaches still suffer from several critical limitations. Implementing feature-level assessment and utilization~\cite{Good features, Illumination-Adaptive, QualiSLAM} can lead to excessive computational complexity. The switching-coupled approach~\cite{switching-coupled} performs independent evaluation and applies uniform assessment strategies across all scenarios, lacking adaptability and continuity. PointSLOT~\cite{PointSlot} manages keyframes separation based solely on dynamic object presence, ignoring cases where background frames may suffer from insufficient quality.

To address these limitations, this paper proposes SR-SLAM, a comprehensive framework that achieves unified scene reliability assessment by integrating multiple environmental metrics and historical evaluation results. Based on this reliability assessment, our approach enables intelligent adaptation across the entire SLAM pipeline through adaptive dynamic processing, reliability-aware pose refinement, and assessment-based new keyframe selection and optimization. This unified reliability-based approach directly addresses the identified limitations while maintaining robust performance across diverse environments.

\section{METHODS}
This section presents the SR-SLAM framework through five key components. Section \uppercase\expandafter{\romannumeral3}.A provides a comprehensive system overview and workflow description. {Section \uppercase\expandafter{\romannumeral3}.B} details the scene reliability assessment mechanism that evaluates environmental reliability for the feature-based method in each frame. {Section \uppercase\expandafter{\romannumeral3}.C} describes the computation of potential dynamic regions and the adaptive dynamic feature culling strategy guided by reliability assessment results. Section \uppercase\expandafter{\romannumeral3}.D presents the pose refinement strategy that leverages direct optimization methods for low-reliability frames. Section \uppercase\expandafter{\romannumeral3}.E introduces reliability-based new keyframe selection and weighted optimization strategies.  
\subsection{Overview of the Proposed System}
\begin{figure*}[!htbp]
  \begin{center}
  \includegraphics[width=\linewidth]{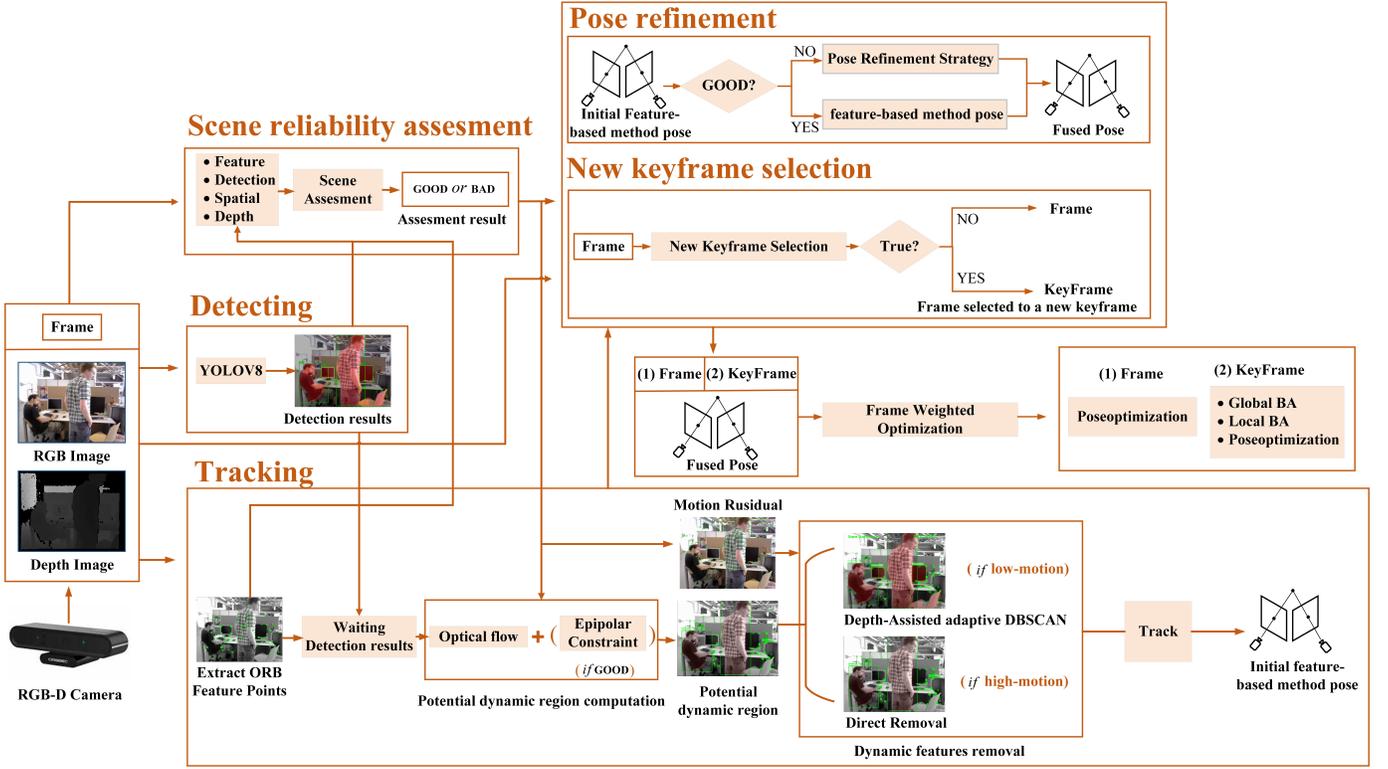}
  \caption{Overview of the SR-SLAM framework. The RGB image is first processed by YOLOv8 for detection, while feature points are extracted from the image. Next, a comprehensive assessment combining RGB, depth information, detection, and features evaluates the scene reliability. Based on this assessment, the system sequentially performs potential dynamic region computation, dynamic feature removal, pose refinement, new keyframe selection, and weighted optimization, completing the entire pipeline.}
  \label{fig:framework}
  \end{center}
\end{figure*}
We focus on feature-based visual SLAM and employ the ORB-SLAM3~\cite{ORB-SLAM3} framework as our foundation. ORB-SLAM3 has been widely adopted and extended by different visual SLAM algorithms to handle complex and dynamic scenarios. Built upon ORB-SLAM3, the SRR-SLAM framework is illustrated in Fig.~\ref{fig:framework}. Under the RGB-D camera model of ORB-SLAM3, we integrate several novel modules: scene reliability assessment, potential dynamic region computation, dynamic feature removal, pose refinement, reliability-based new keyframe selection, and weighted optimization into the existing framework while introducing an additional detecting thread.

When SRR-SLAM operates, RGB-D frames (comprising RGB and depth images) from the camera are simultaneously distributed to three components: the scene reliability assessment module, the detection thread, and the tracking thread. The scene reliability assessment module directly receives depth images to compute depth quality metric. The tracking thread performs ORB feature extraction on the input RGB images, and the extracted ORB feature points are forwarded to the scene reliability assessment module to derive feature quality metric. Simultaneously, the detecting thread processes the input RGB images using YOLOv8~\cite{Yolov8} to generate detection results and identify \textit{a priori} dynamic objects. The detection results from the detecting thread are then sent to both the tracking thread and the scene reliability assessment module to compute detection confidence and object spatial metrics.

Based on these comprehensive metrics, the scene reliability assessment module evaluates the reliability for the feature-based method, determining whether the current frame should be classified as \textbf{GOOD} or \textbf{BAD}, indicating whether it is suitable or unsuitable for applying the feature-based method. This reliability assessment is then distributed to potential dynamic region computation, dynamic feature removal, pose refinement, new keyframe selection, and weighted optimization modules.

Upon receiving the reliability result, the tracking thread performs potential dynamic region computation using either the Lucas-Kanade (LK) optical flow method alone or combined with epipolar constraints to match compute the potential dynamic area. During the subsequent dynamic feature culling process, the motion information from potential dynamic region computation guides the removal strategy: either direct removal for high-motion scenarios or depth-assisted adaptive DBSCAN for low-motion cases. The remaining feature points are then used for tracking to compute the initial camera pose via the feature-based method.

For low-reliability scenes, the initial feature-based camera pose undergoes pose refinement in the \textbf{TrackLocalMap} process, where direct methods generate a compensated fused pose with multiple historical good frames as reference. For high-reliability scenarios, the original initial feature-based pose is directly used as the fused pose.

After completing the above processes, scene reliability guides new keyframe selection through threshold adjustment in the \textbf{NeedNewKeyFrame} function, ensuring that frames participating in subsequent optimization maintain high reliability. To handle consecutive low-reliability scenarios, we implement a threshold reduction mechanism to maintain new keyframe selection continuity. During optimization, since the above strategy may introduce low-reliability keyframes, we apply information matrix weight penalties to low-reliability keyframes and also apply them to low-reliability frames, allowing the optimization processes, including \textbf{BundleAdjustment} (BA) and \textbf{PoseOptimization} to prioritize high-reliability observations.

\subsection{Scene Reliability Assessment Mechanism}
In the scene reliability assessment process, we first perform 
an evaluation on the initial frames (5 frames in our method), selecting the frame with the highest reliability as the preliminary reference frame to complete the initialization process. For subsequent frame processing, we 
obtain the degree of change (motion residual) between the current frame and the reference frame, then use the same evaluation method to obtain the 
reliability of the current frame (if it is higher than the previous reference frame, we update the reference frame information). Finally, we use the degree of change as a weight to process the current evaluation reliability, ultimately obtaining a comprehensive reliability score.

For current frame evaluation, we are inspired by the objects classification criteria in~\cite{switching-coupled} and the challenge evaluation module~\cite{QualiSLAM, CEMS, visual attention model, Overcoming occlusion, CG-SLAM} to establish our overall reliability assessment calculation method and select four complementary metrics to design a scene reliability assessment mechanism consisting of: Object detection confidence and Spatial distribution, Feature Quality, and Depth Quality. This module aims to estimate how suitable the current scene is for feature-based SLAM methods. The overall reliability $R$ is calculated as:
\begin{equation}
R_c = \sum_{i=1}^{4} R_i \quad i \in \{conf, spatial, feature, depth\}
\label{eq:current reliability}
\end{equation}
where $R_c$ represents the current reliability of the frame, and $R_i$ represents the $i$-th metric, corresponding to the four metrics of Object detection confidence and Spatial distribution, Feature Quality mentioned above.

The first metric follows the design of the observation quality metric in~\cite{switching-coupled} to reflect the accuracy of object detection under current conditions, which is crucial for the effectiveness of subsequent dynamic region selection based on detection results. This metric computes the average detection confidence of all objects detected by YOLOv8 in the current frame as

\begin{equation}
R_{conf} = \frac{1}{N}\sum_{i=1}^{N}conf_i
\label{eq:confidence}
\end{equation}
where $N$ is the number of detected objects and $conf_i$ is the confidence score of the $i$-th object in the current frame.

The second metric, Spatial Distribution, draws from the role of object layout in visual attention modeling~\cite{visual attention model} and the impact of object occlusion~\cite{Overcoming occlusion} in autonomous driving scenes. Accordingly, this metric combines object distribution and size characteristics to provide a comprehensive evaluation that reflects both visual attention effects and occlusion conditions of objects in the current scene as
\begin{equation}
\begin{split}
R_{spatial} &= \frac{1}{N}\sum_{i=1}^{N}\{\max\left(0,1-\frac{A_i}{A_t}\right) 
 + \left(1-\min\left(1,\frac{d_i}{d_m}\right)\right)\}
\label{eq:spatial}
\end{split}
\end{equation}
where $N$ is the number of detected objects, $A_i$ is the area of the $i$-th object, $A_t$ is the total frame area, $d_i$ is the distance from the object's center to the image center, and $d_m$ is the image diagonal length.

The third metric, Feature Quality, extends the idea of evaluating feature properties on individual objects (as in \cite{switching-coupled}) to the entire scene. Following the grid-based image representation techniques proposed in~\cite{QualiSLAM} and the requirements for feature distribution and feature richness in geometric constraints~\cite{Multiple View Geometry} and feature-based methods, we divide the image into $G$ (3×3) grids and assess the feature distribution and richness within each cell to reflect the suitability of current scene features for subsequent geometric methods and tracking processes, as
\begin{equation}
R_{feature} = \frac{1}{G}\sum_{i=1}^{G}\left\{\frac{\sum_{p=1}^{c_i}r_p}{c_i} + \frac{1}{1+\sqrt{\sigma_i^2}} \right\}
\label{eq:feature}
\end{equation}
where $G$ represents the number of grids. In the $i$-th grid, $r_p$ represents the feature response value of the $p$-th feature point, $c_i$ represents the feature count, and $\sigma_i^2$ is the variance of the feature distribution.

Finally, the fourth metric, Depth Quality, also employs the grid-based method used in Feature Quality assessment. It combines three key components: depth distribution entropy, spatial coverage, and scene structure consistency (inspired by the depth uncertainty modeling in~\cite{CG-SLAM}) to produce a comprehensive depth quality metric that indicates whether depth information is reliable for supporting feature matching and motion estimation, as
\begin{equation}
\begin{split}
R_{depth} &= \frac{1}{G}\sum_{i=1}^{G}\left\{\frac{N_{v,i}}{N_{t,i}} 
+ \left(1-\min\left(1,\frac{\sigma_{d,i}}{C_m}\right)\right) \right.\\
&\quad + \left.\left(1-\min\left(1,\frac{\mu_{g,i}}{S_m}\right)\right)\right\}
\label{eq:depth}
\end{split}
\end{equation}
where $G$ represents the number of grids. In the $i$-th grid, $N_{v,i}$ and $N_{t,i}$ are the valid and total depth pixel counts respectively, $\sigma_{d,i}$ is the depth standard deviation, and $\mu_{g,i}$ is the depth gradient mean, with $C_m$ and $S_m$ representing maximum thresholds of $\sigma_{d,i}$ and $\mu_{g,i}$ to prevent abnormal depth variations.

Following the computation of current reliability $R_c$, to better evaluate the current reliability in terms of continuity and historical consistency, we determine the motion residual as the degree of change between the current frame and the corresponding reference frame through optical flow analysis combined with the grid-based method, as 
\begin{equation}
{R_r} = \frac{{\sum\nolimits_{i = 1}^N {\left\| {{m_i}} \right\|} }}{N} + \frac{{\sum\nolimits_{j = 1}^{{G}} {\left\| {{M_j}} \right\|} }}{{{G}}}
\label{eq:Motion Residual}
\end{equation}
where $m_i$ represents the point-level motion magnitude for the $i$-th feature point, $M_j$ denotes the grid-level motion for the $j$-th grid, $N$ is the total number of feature points, and $G$ is the number of grids.

Finally, we conduct a weighted fusion of the degree of change ($R_r$) and the current reliability ($R_c$) to derive the final scene reliability for the current frame. Therefore, in highly dynamic and complex environments, the final reliability undergoes additional reduction beyond the original current reliability assessment. Through these evaluation procedures, we accomplish a comprehensive assessment of scene reliability for feature-based methods. Subsequently, we compare $R$ with $th_{scene}$ to obtain the binary classification result of \textbf{GOOD} or \textbf{BAD} for the current frame. During the initialization process, the degree of change component is excluded from the calculation due to the unavailability of reference frames.
\begin{equation}
R = \frac{R_r}{R_{cb}} \cdot R_c
\label{eq:final_reliability}
\end{equation}

\begin{equation}
\text{Scene} = \begin{cases}
\textbf{GOOD} & \text{if } R \geq th_{scene} \\
\textbf{BAD} & \text{otherwise}
\end{cases} 
\label{eq:final_decision}
\end{equation}
where $R_{cb}$ serves as a change baseline for measuring the degree of change and the $th_{scene}$ serves as the classification threshold. Both $R_{cb}$ and $th_{scene}$  are derived from the same relatively stable datasets (fr3/sitting/static in TUM and synchronous1,2 in BONN). We set $R_{cb}$ to the maximum value (60 in our method) from the observed motion residual distribution to establish an upper bound for the degree of change, while $th_{scene}$ is set to the minimum reliability value (0.20 in our method) to ensure conservative classification. 

Through evaluation procedures based on multiple metrics and historical reference information, our approach ensures both the continuity of the assessment process and its adaptability across diverse environments, while accomplishing comprehensive scene reliability assessment for feature-based methods and obtaining the corresponding \textbf{GOOD}/\textbf{BAD} classification. This reliability assessment subsequently guides our dynamic culling strategies, new keyframe selection, and optimization processes.

\subsection{Dynamic Culling Strategies}

As illustrated in Algorithm~\ref{alg:dynamic_culling}, our dynamic culling strategy operates in two main stages: potential dynamic region computation and adaptive dynamic feature removal after the scene reliability has been evaluated.

\begin{algorithm}[htbp]
\caption{Dynamic Culling Strategies}
\label{alg:dynamic_culling}
\begin{algorithmic}[1]
\REQUIRE All feature Points, Scene reliability $R$, feature reliability $R_{feature}$, Motion residual $R_r$, detection results.
\ENSURE Static feature set for tracking

\STATE \textbf{Stage 1: Potential Dynamic Region Computation}
\STATE Apply LK optical flow to extract anomalous features
\IF{$R$ classified as \textbf{GOOD}}
    \STATE Add epipolar constraints 
\ELSE
    \STATE Use only LK optical flow
\ENDIF
\STATE Combine geometric anomalies with object detection results
\STATE Define potential dynamic regions

\STATE \textbf{Stage 2: Adaptive Dynamic Feature Removal}
\IF{Initialization phase}
    \STATE Apply depth-assisted adaptive DBSCAN (stable assumption)
\ELSE
    \IF{$R_r > 60$ (High-motion)}
        \STATE Apply aggressive removal (remove all potential dynamic points)
    \ELSE
        \STATE Apply depth-assisted adaptive DBSCAN
        \STATE \quad \textbf{Depth-based pre-clustering}
        \STATE \quad \quad  Compute $\epsilon_{depth}$ using IQR of dynamic points
        \STATE \quad \quad  Group points by depth similarity
        \STATE \quad \textbf{Adaptive DBSCAN within each depth cluster}
        \STATE \quad \quad  Adapt $eps$ based on scene reliability $R$
        \STATE \quad \quad  Adapt $minPts$ based on feature reliability 
        \STATE \quad \quad $R_{feature}$
        \STATE \quad \quad Apply DBSCAN clustering
    \ENDIF
\ENDIF

\RETURN Static features 
\end{algorithmic}
\end{algorithm}

In the potential dynamic region computation stage, we first apply LK optical flow to all frames for preliminary anomalous feature detection. Based on our reliability assessment, if classified as \textbf{GOOD}, the scene indicates stable detection with sufficient feature quantity and distribution quality for reliable epipolar constraint computation. In this case, we incorporate epipolar constraints; otherwise, we use only LK optical flow. The resulting geometrically anomalous features are combined with object detection results to define potential dynamic regions, which include all features within predefined dynamic object regions and regions containing geometrically anomalous points outside objects.

In the subsequent dynamic feature removal stage, the computed motion information guides the selection between aggressive or refined removal strategies. We directly utilize the degree of change $R_r$ (motion residual) obtained from Section \uppercase\expandafter{\romannumeral3}.B to distinguish between low-motion and high-motion scenarios, with the same threshold (60) for consistency. 

\begin{figure*}[!htbp]
  \begin{center}
  \includegraphics[width=\linewidth]{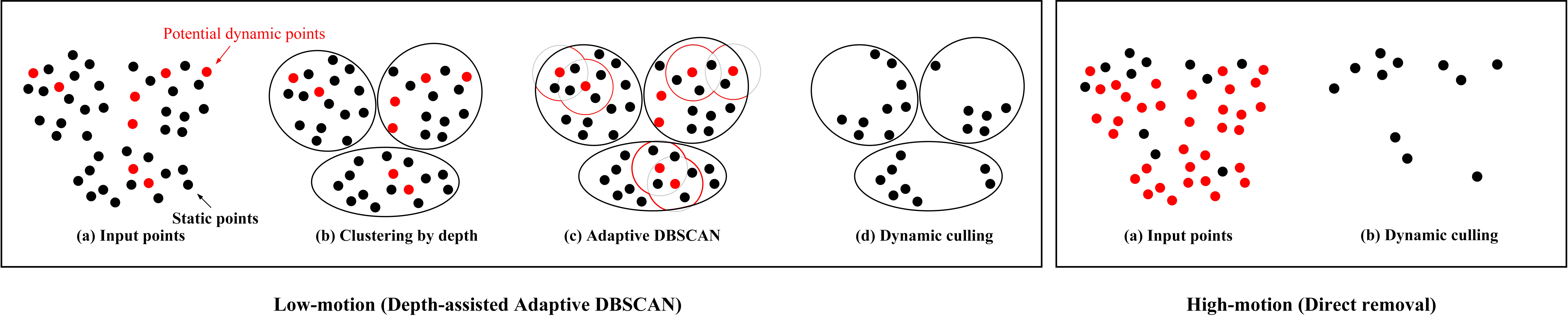}
  \caption{Dynamic Culling Strategy of SR-SLAM system. Black points represent static point features, and} red points represent potential dynamic point features. Black circles indicate depth-based clustering results, and  circles denote the final dynamic regions identified by the depth-assisted adaptive DBSCAN.
  \label{fig:Dynacmic Culling}
  \end{center}
\end{figure*}

As shown in Fig.~\ref{fig:Dynacmic Culling}, when the current scene is high-motion, where dynamic features dominate, aggressive removal of all potential dynamic points is adopted since the harm of erroneously retaining dynamic features far outweighs the benefit of preserving additional static features. In contrast, in low-motion scenarios, where dynamic features are sparse, we employ our proposed depth-assisted adaptive DBSCAN for refined processing of all geometrically anomalous features, aiming to preserve maximum usable static features while maintaining precision. During the initialization process, since reference frames are unavailable for motion residual computation, we directly employ depth-assisted adaptive DBSCAN for robust initialization, following the environmental stability assumption for initialization processes in SLAM systems~\cite{ORB-SLAM2,ORB-SLAM3}.

In our depth-assisted adaptive DBSCAN, we first perform depth-based pre-clustering to group points with similar depth characteristics, enabling subsequent DBSCAN to operate more effectively within each depth cluster for refined dynamic object removal. We determine the depth clustering threshold by computing the interquartile range (IQR) of all potential dynamic features, which effectively captures their distribution characteristics while being robust to extreme depth values. This allows dynamic features to be accurately and efficiently grouped during the clustering process. The calculations are as follows
\begin{align}
Q1(D_{depth}) &= D_{depth}[\lfloor 0.25 \times n \rfloor] \\
Q3(D_{depth}) &= D_{depth}[\lfloor 0.75 \times n \rfloor]
\end{align}
\begin{equation}
\epsilon_{depth} = Q3(D_{depth}) - Q1(D_{depth})
\label{eq:depth threshold}
\end{equation}
where $D_{depth}$ represents the sorted depth values of all potential dynamic points, $n$ is the total number of potential dynamic points, $Q1$ is the 25th percentile, $Q3$ is the 75th percentile of the $D_{depth}$ and $\epsilon_{depth}$ is the depth clustering threshold.

After completing the depth clustering, we apply our adaptive DBSCAN algorithm. DBSCAN requires two critical parameters: $eps$ (maximum search range) and $minPts$ (minimum number of points for cluster formation). We establish parameter base values following~\cite{Object-oriented}, setting $eps_{base} = 0.02$ and $minPts_{base} = 3$. To achieve adaptive adjustment, we adjust the maximum search range $eps$ based on overall scene reliability $R$, expanding the search range in high-reliability scenarios. For $minPts$, we use feature reliability $R_{feature}$ to accommodate more features per cluster when features are abundant and well-distributed. The adaptive calculation employs sigmoid functions for smooth parameter modulation as follows:
\begin{equation}
\text{$ad$}(R, \alpha) = \frac{\text{$Minscale$} + \text{$Maxscale$}}{1 + e^{-\alpha \cdot R_i}}
\label{eq:adapt_param}
\end{equation}
\begin{equation}
eps = eps_{base} \cdot \text{$ad$}(R, \alpha_{eps})
\label{eq:epsilon}
\end{equation}
\begin{equation}
\text{$minPts$} = \text{$minPts$}_{base} \cdot \text{$ad$}(R_{feature}, \alpha_{min})
\label{eq:minpts}
\end{equation}
where $\text{$ad$}(\cdot)$ is the adaptive DBSCAN function, $R$ and $R_{feature}$ represent scene and feature reliability respectively, $\alpha_{eps}$ and $\alpha_{min}$ control the adaptation sensitivity, $R_i$ can refer to either $R$ or $R_{feature}$, and $\text{$MinScale$}$, $\text{$Maxscale$}$ define the parameter adjustment bounds (set with 0.5 and 2.0 in our method).

With our depth-assisted adaptive DBSCAN approach, we achieve precise dynamic point identification and removal while reducing computational complexity from $\mathcal{O}(n^2)$ to $\mathcal{O}(n^2/N)$ with high-dimensional data. As shown in Fig.~\ref{fig:removal_compare}, we compared different approaches (Proportion, K-means, Convex Hull, standard DBSCAN, and our proposed depth-assisted adaptive DBSCAN) under low-dynamic and moderate-dynamic conditions. 

In the low-dynamic scenario, where the two individuals exhibit occasional head and upper body movements with subtle gestures, the Proportion, K-means, and Convex Hull methods retain certain static features, such as the lower body of the person on the right and points on the chair. However, due to the rigid cluster shapes of K-means and Convex Hull and the limitations of proportion-based filtering, these methods struggle with irregular shapes such as a seated person in an L-shaped pose. Specifically, the upper-body surrounding features of the person on the left are entirely removed, and the upper-right region around the person on the right is also over-filtered. In contrast, the standard DBSCAN method retains more useful features due to its flexible cluster boundaries, preserving both the surrounding features near the person on the left and the upper-right region of the person on the right. However, lacking depth awareness, it erroneously retains dynamic points on the head of the person on the right and the upper body of the person on the left, particularly near the monitor area. Our depth-assisted adaptive DBSCAN addresses these shortcomings by leveraging depth information and self-adjusting parameters. As a result, it successfully preserves nearby static features while accurately removing the dynamic points on the upper bodies and heads of both individuals. 

In the moderate-dynamic scenario, where more pronounced body movement is present, the limitations of Proportion, K-means, and Convex Hull become more evident. These methods fail to adapt to complex human shapes and remove not only dynamic features but also many valuable surrounding static features. Standard DBSCAN performs better, retaining more useful features. However, it still leaves many residual dynamic points, especially on the left arm of the individual. In comparison, our depth-assisted adaptive DBSCAN not only removes the dynamic features correctly, including the challenging arm area, but also avoids over-filtering, demonstrating superior adaptability and robustness in dynamic environments.

By leveraging scene assessment, our dynamic culling strategies both ensure that geometric constraints align with the current environment, while also improving the adaptability of removal algorithms and the handling of high-dimensional data across diverse environments, achieving excellent dynamic removal, feature retention, and computational efficiency across diverse scenarios.
\begin{figure*}[!htbp]
  \begin{center}
  \includegraphics[width=\linewidth]{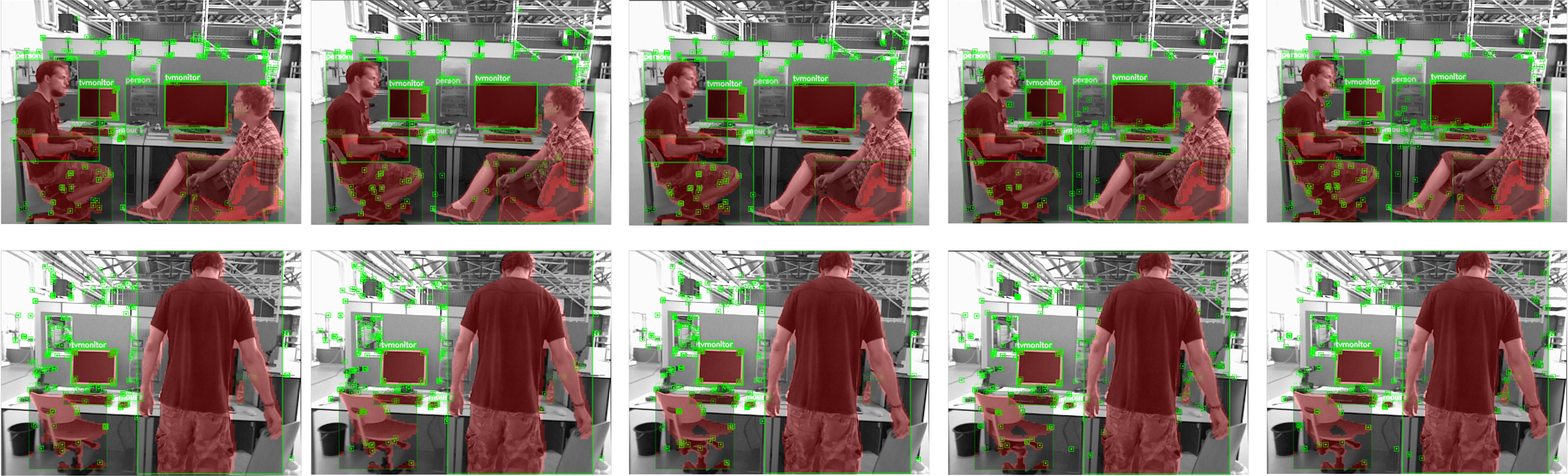}
  \caption{Comparison of feature removal methods under different dynamic conditions. The first row shows results in a low dynamic scene, while the second row corresponds to a moderately dynamic environment. From left to right: proportion-based method, K-means clustering, ConvexHull-based method, standard DBSCAN, and our proposed depth-assisted adaptive DBSCAN.}
  \label{fig:removal compare}
  \end{center}
\end{figure*}

\subsection{Pose Refinement}
When a frame is classified as BAD, indicating poor reliability for feature-based methods, we employ a direct method to refine the pose estimation. Drawing inspiration from RigidFusion\cite{RigidFusion} and StaticFusion\cite{StaticFusion}, a direct method 
uses motion residuals for rapid static/dynamic frame classification and 
optimization for pose estimation to maintain assessment continuity. Our approach computes relative poses between the current BAD frame and the $K$ temporally closest GOOD frames, where $K$ represents the number of reference frames retained for pose refinement (empirically set to 5 in our implementation). The direct method optimization employs photometric and geometric residuals. The photometric residual measures intensity consistency between warped and actual pixel values:
\begin{equation}
r_I^p = I_B(W(x_A^p, T(\xi), D_A)) - I_A(x_A^p)
\label{eq:intensity_residual}
\end{equation}
The geometric residual ensures depth consistency after transformation:
\begin{equation}
r_D^p = D_B(W(x_A^p, T(\xi), D_A)) - |T(\xi)\pi^{-1}(x_A^p, D_A(x_A^p))|_z
\label{eq:depth_residual}
\end{equation}
where $I_A$ and $I_B$ are RGB images, $D_A$ and $D_B$ are depth images, $W$ is the warping function that projects pixels from frame A to frame B based on transformation $T(\xi)$, $\pi^{-1}$ represents inverse projection from 2D to 3D space, $x_A^p$ is the 2D coordinate of pixel $p$ in frame A, and $|\cdot|_z$ denotes the z-component of the transformed 3D point. The combined cost function integrates both residual types with appropriate weighting:
\begin{equation}
\mathcal{D} = C(\alpha_I w_I^p r_I^p(\xi)) + C(w_D^p r_D^p(\xi))
\label{eq:combined_cost}
\end{equation}
where $w_I^p$ and $w_D^p$ are pixel-specific weights computed according to photometric and depth measurement noise, $C(\cdot)$ is a robust cost function as in StaticFusion\cite{StaticFusion}, and $\alpha_I$ is a scale parameter ensuring photometric and depth residuals are comparable. The final optimization objective incorporates temporal weighting and dynamic region handling:
\begin{equation}
\mathcal{R}(\xi) = \sum_{p=1}^{N} \sum_{k=1}^{K} B_k \cdot \gamma_i(p) \cdot \mathcal{D}_k(p,\xi)
\label{eq:direct_optimization}
\end{equation}
where $\gamma_i(p)$ represents pixel-specific weights assigning lower values to foreground and higher values to static background pixels, $B_k=e^{-\lambda(t_k-t_c)}$ is a time-decay factor prioritizing more recent reference frames with $\lambda$ controlling the temporal decay rate, $k$ indexes the reference frames, and $\mathcal{D}_k(p,\xi)$ represents the combined cost from Eq.~\ref{eq:combined_cost} for frame $k$.
The final pose combines feature-based and direct method estimates through adaptive weighting:
\begin{equation}
T_{fuse} = w_{diff} \cdot T_{feature} + (1-w_{diff}) \cdot T_{direct}
\label{eq:optimized_pose}
\end{equation}
where $w_{diff} = e^{-\mu \cdot R_{r}}$ adapts the weighting based on the degree of change $R_{r}$ (motion residual) in Section \uppercase\expandafter{\romannumeral3}.B, $\mu$ controls sensitivity to the degree of change, $T_{feature}$ represents the pose from feature-based tracking, and $T_{direct}$ is the direct method result. This adaptive mechanism ensures smooth transitions between these two estimation methods.

Through the above direct method compensation using historical reference information for low-reliability feature-based situations, we address pose estimation deficiencies when current features are insufficient for feature-based methods. This approach enables our system to achieve more stable pose estimation and robust performance across various scenarios that lead to feature insufficiency.

\subsection{New Keyframe Selection and Weighted Optimization}

Following the processes above, we implement an intelligent new keyframe selection mechanism to ensure high-quality keyframes while minimizing interference from unstable factors during BA (bundle adjustment) optimization. Inspired by variance-based selection methodologies in QualiSLAM~\cite{QualiSLAM} and feature selection principles~\cite{feature selection}, where high-variance metrics containing rich information are retained while low-variance metrics are discarded, we apply a similar principle to new keyframe selection.

In our context, the variance of BAD frames indicates their likelihood of approaching GOOD frame quality; high variance suggests greater potential for high-reliability, while low variance indicates consistently poor conditions with low-reliability. Our selection mechanism ideally ensures stable keyframe conditions by preferring GOOD frames, however, in practical scenarios such as continuous BAD frame sequences, maintaining strict quality requirements would lead to keyframe starvation and compromise system continuity.

Therefore, we implement adaptive threshold adjustment that dynamically relaxes the selection criteria when necessary, ensuring sufficient keyframes for robust BA optimization while preserving overall system stability. For GOOD frames ($R$ higher than $th_{scene}$), we directly accept them as keyframes. For BAD frames, we propose a dual-constraint approach based on the current frame's reliability $R_{bad}$, dynamically adjusting the new keyframe selection threshold through two complementary components: threshold degradation allowance (for continuity) and quality preservation constraint (to prevent introduction of excessively poor frames).

We first compute historical statistics to establish the foundation for adaptive adjustment:
\begin{equation}
\bar{R}_{bad} = \frac{1}{|W|}\sum_{i \in W} R_i, \quad \sigma^2_{bad} = \frac{1}{|W|}\sum_{i \in W} (R_i - \bar{R}_{bad})^2
\label{eq:statistics}
\end{equation}
where $W$ represents the sliding window of recent BAD frames (reliability lower than $th_{scene}$, window size = 20), $R_i$ is the $i$-th BAD frame. $\bar{R}_{bad}$ is the mean reliability of historical BAD frames, and $\sigma^2_{bad}$ is their variance, indicating the stability of poor-quality conditions.

Subsequently, we compute the variance-driven weight that balances the two constraint components:
\begin{equation}
w_{select} = w \cdot \min\left(\frac{\sigma^2_{bad}}{\gamma}, 1\right)
\label{eq:variance_weight}
\end{equation}
where $w$ is the base value of the weight (set to 1.0), and $w_{select}$ determines the relative importance of degradation allowance versus preservation constraint. High variance yields a higher weight for threshold relaxation (enabling continuity), while low variance maintains stricter quality requirements. The parameter $\gamma$ serves as the reference variance value, and the weight increases linearly with variance until reaching the maximum weight.

The adaptive threshold integrates both constraints through the variance-weighted combination:
\begin{equation}
\begin{split}
th_{adaptive} = th_{scene} + w_{select} \cdot (R_{bad} - \bar{R}_{bad}) \\
\quad - (1 - w_{select}) \cdot (R_{bad} - \bar{R}_{keyframe})
\end{split}
\label{eq:adaptive_threshold}
\end{equation}
where the first term provides degradation allowance based on the current frame's mean reliability from historical BAD frames, while the second term enforces preservation constraint based on mean relibility from historical keyframe quality baseline $\bar{R}_{keyframe}$, and $th_{scene}$ is the threshold distinguishing GOOD and BAD frames. Finally, new keyframe selection for BAD frames:
\begin{equation}
\text{New KeyFrame} = \begin{cases}
\text{True} & \text{if} \hspace{1mm} R_{bad} \geq th_{adaptive} \\
\text{False}  & \text{otherwise}
\end{cases}
\label{eq:keyframe_selection}
\end{equation}

This mechanism ensures that new keyframe selection maintains continuity during challenging periods while preserving overall quality standards. 

After new keyframe selection, since BAD frames may be introduced through the adaptive mechanism, we implement a frame-level penalty mechanism that applies information matrix weighting to all current BAD frame features during both keyframe BA optimization and individual frame pose optimization processes.
\begin{equation}
Q_{bad}^{-1} = \frac{R_{bad}}{th_{scene}} \cdot Q^{-1}
\label{eq:information_matrix}
\end{equation}
where $Q^{-1}$ is the original information matrix, $R_{bad}$ is the reliability of the current BAD frame, and $Q_{bad}^{-1}$ is the weighted information matrix for the current BAD frame. By comparing the current reliability against the classification threshold, we penalize the information matrix to reduce the influence of all features in BAD frames during the optimization process, ensuring that lower-quality frames contribute proportionally less to pose estimation.

Through the integration of intelligent new keyframe selection with variance-weighted strategy and frame-level information matrix penalty mechanisms, we ensure stable optimization inputs while maintaining continuous system operation and adjusting frame contributions based on their reliability in optimization. This approach resolves feature-level operational complexity while considering real-world conditions, making optimization more precise and robust across diverse scenarios.

\section{EXPERIMENT AND RESULTS}
To comprehensively evaluate and validate our SR-SLAM method, we conducted extensive experiments on two public RGB-D datasets (TUM\cite{TUM} and BONN\cite{BONN}) as well as real-world environments. We employ absolute trajectory error (ATE) and relative pose error (RPE) as primary evaluation metrics for trajectory accuracy assessment. The root mean square error (RMSE) and standard deviation (S.D.) are utilized to characterize both trajectory accuracy and system stability. The RPE evaluation encompasses relative translation error (T.RPE) and relative rotation error (ROT.RPE) to provide comprehensive pose estimation analysis.

For dataset-based evaluation, we validate the performance improvements of our method relative to ORB-SLAM3~\cite{ORB-SLAM3} as the baseline. Additionally, we conduct comparative analysis with state-of-the-art dynamic SLAM methods based on ORB-SLAM2~\cite{ORB-SLAM2} and ORB-SLAM3~\cite{ORB-SLAM3} frameworks developed in recent years. In the result tables, the best result for each sequence is highlighted in bold, while the second best result is underlined for clarity. And all improvements of SR-SLAM against ORB-SLAM3 is calculated by:
\begin{equation}
P_{im}= (1 - \frac{V_{SR}}{V_{ORB}} \cdot 100\%)
\label{eq:improvements}
\end{equation}
where $P_{im}$ represents the improvement percentage, $V_{SR}$ and $V_{ORB}$ represent values of SR-SLAM and ORB-SLAM3, respectively.

For real-world validation, we conducted experiments using both fixed camera configurations to verify performance across various scenarios, and mobile robot platforms to assess robustness during moving in various scenarios. All experiments were conducted on a computer equipped with Ubuntu 20.04 operating system, AMD Ryzen R7 CPU, NVIDIA RTX3080 GPU, and 32 GB of RAM. An Intel RealSense D455 camera was employed for both fixed camera and mobile robot experimental setups to ensure consistent data acquisition conditions.

\subsection{Evaluation on the TUM RGB-D Dataset}
The TUM RGB-D dataset, developed by the Technical University of Munich Computer Vision Group, provides a comprehensive benchmark for visual odometry and SLAM systems. The dataset encompasses diverse camera motion patterns, including hemispherical, XYZ translation, RPY rotation, and static configurations across various indoor scenarios.

To thoroughly validate our method across different challenging conditions, we selected five representative sequences: fr3/walking/half, fr3/walking/xyz, fr3/walking/rpy, fr3/walking/static, and fr3/sitting/static. The first three sequences feature active camera motion combined with frequent pedestrian movement involving two individuals, creating complex dynamic environments. The fr3/walking/static sequence presents a static camera configuration with minimal human motion, while fr3/sitting/static depicts seated individuals with only occasional hand gestures and head movements, representing subtle dynamic interference scenarios.

For quantitative evaluation, we employed the aforementioned metrics: RMSE and S.D. of ATE to measure global trajectory consistency, T.RPE to assess translational drift error, and ROT.RPE to evaluate rotational drift accuracy. We conducted a comparative analysis in two phases: first, comparing our method against ORB-SLAM2 and four state-of-the-art dynamic SLAM methods based on ORB-SLAM2, followed by evaluation against ORB-SLAM3 and three state-of-the-art dynamic SLAM methods built upon ORB-SLAM3. The comprehensive results are presented in Tables~\ref{tab:comprehensive_results based on orbslam2} and~\ref{tab:comprehensive_results based on orbslam3}.

In Table~\ref{tab:comprehensive_results based on orbslam2}, we compare ORB-SLAM2 with four state-of-the-art dynamic SLAM methods: Blitz-SLAM~\cite{BlitzSLAM}, YOLO-SLAM~\cite{YOLO-SLAM}, SG-SLAM~\cite{SG-SLAM}, and SSF-SLAM~\cite{SSF-SLAM}. Our SR-SLAM achieves nearly 90\% improvements in ATE and RPE over ORB-SLAM2 on dynamic sequences while maintaining up to 50\% accuracy gains on low-dynamic sequences.
\begin{table*}[!htbp]
\centering
\caption{COMPARATIVE RESULTS OF DIFFERENT SLAM SYSTEMS BASED ON ORB-SLAM2.}
\label{tab:comprehensive_results based on orbslam2}
\small  
\vspace{-2mm}
\setlength{\tabcolsep}{8pt}
\renewcommand{\arraystretch}{1.3}
\begin{tabular}{l|c@{\hspace{8pt}}c|c@{\hspace{8pt}}c|c@{\hspace{8pt}}c|c@{\hspace{8pt}}c|c@{\hspace{8pt}}c|c@{\hspace{8pt}}c}
\hline
\rule{0pt}{3ex}
& \multicolumn{2}{c|}{ORB-SLAM2} & \multicolumn{2}{c|}{Blitz-SLAM} & \multicolumn{2}{c|}{YOLO-SLAM} & \multicolumn{2}{c|}{SG-SLAM} & \multicolumn{2}{c|}{SSF-SLAM} & \multicolumn{2}{c}{OURS} \\
\rule{0pt}{3ex} 
Sequences & RMSE & S.D. & RMSE & S.D. & RMSE & S.D. & RMSE & S.D. & RMSE & S.D. & RMSE & S.D. \\
\hline
\multicolumn{13}{c}{\textbf{ABSOLUTE TRAJECTORY ERROR (ATE)}} \\
\hline
\rule{0pt}{3ex} 
\hspace{-1mm}fr3/w/half & 0.5176 & 0.2342 & \underline{0.0256} & \underline{0.0126} & 0.0283 & 0.0138 & 0.0268 & 0.0134 & 0.0264 & 0.0140 & \textbf{0.0242} & \textbf{0.0120} \\[2pt]
fr3/w/xyz & 0.8093 & 0.4719 & 0.0153 & 0.0078 & \textbf{0.0146} & \underline{0.0070} & 0.0152 & 0.0075 & \underline{0.0148} & \textbf{0.0067} & 0.0150 & 0.0075 \\[2pt]
fr3/w/rpy & 1.0466 & 0.5669 & 0.0356 & 0.0220 & 0.2164 & 0.1001 & 0.0324 & 0.0187 & \textbf{0.0211} & \textbf{0.0097} & \underline{0.0278} & \underline{0.0145} \\[2pt]
fr3/w/static & 0.4103 & 0.2036 & 0.0102 & 0.0052 & \underline{0.0073} & \textbf{0.0034} & \underline{0.0073} & \textbf{0.0034} & \textbf{0.0070} & 0.0038 & 0.0091 & \underline{0.0037} \\[2pt]
fr3/s/static & 0.0082 & 0.0040 & -- & -- & 0.0066 & \textbf{0.0029} & \underline{0.0060} & \textbf{0.0029} & 0.0067 & 0.0031 & \textbf{0.0055} & \textbf{0.0029} \\[2pt] 
\hline
\multicolumn{13}{c}{\textbf{METRIC TRANSLATIONAL DRIFT (T.RPE)}} \\
\hline
\rule{0pt}{3ex} 
\hspace{-1mm}fr3/w/half & 0.3533 & 0.2832 & \underline{0.0253} & \underline{0.0123} & 0.0268 & 0.0124 & 0.0279 & 0.0146 & 0.0280 & 0.0142 & \textbf{0.0129} & \textbf{0.0073} \\[2pt]
fr3/w/xyz & 0.4356 & 0.2913 & 0.0197 & 0.0096 & 0.0194 & 0.0097 & 0.0194 & 0.0100 & \underline{0.0171} & \textbf{0.0073} & \textbf{0.0131} & \underline{0.0078} \\[2pt]
fr3/w/rpy & 0.4044 & 0.2923 & 0.0473 & 0.0283 & 0.0933 & 0.0736 & 0.0450 & 0.0262 & \underline{0.0327} & \underline{0.0179} & \textbf{0.0194} & \textbf{0.0119} \\[2pt]
fr3/w/static & 0.2021 & 0.1704 & 0.0129 & 0.0069 & 0.0094 & 0.0044 & 0.0100 & \underline{0.0051} & \underline{0.0093} & \underline{0.0051} & \textbf{0.0074} & \textbf{0.0043} \\[2pt]
fr3/s/static & 0.0089 & 0.0045 & -- & -- & 0.0089 & 0.0044 & \underline{0.0075} & \underline{0.0035} & 0.0079 & \underline{0.0035} & \textbf{0.0049} & \textbf{0.0025} \\[2pt]
\hline
\multicolumn{13}{c}{\textbf{METRIC ROTATIONAL DRIFT (ROT. RPE)}} \\
\hline
\rule{0pt}{3ex} 
\hspace{-1mm}fr3/w/half & 7.2536 & 5.7610 & 0.7879 & 0.3751 & \underline{0.7534} & \underline{0.3564} & 0.8119 & 0.3878 & 0.7966 & 0.4199 & \textbf{0.4061} & \textbf{0.2163} \\[2pt]
fr3/w/xyz & 8.4536 & 5.7334 & 0.6132 & 0.3348 & 0.5984 & 0.3655 & 0.5040 & \underline{0.2469} & \textbf{0.3912} & \textbf{0.1610} & \underline{0.3939} & 0.2752 \\[2pt]
fr3/w/rpy & 7.9027 & 5.6830 & 1.0841 & 0.6668 & 1.8238 & 1.4611 & 0.9565 & 0.5487 & \underline{0.7395} & \textbf{0.4013} & \textbf{0.6151} & \underline{0.4693} \\[2pt]
fr3/w/static & 3.4117 & 2.8485 & 0.3038 & 0.1437 & 0.2676 & 0.1104 & 0.2678 & \underline{0.1144} & \underline{0.2473} & 0.1153 & \textbf{0.1900} & \textbf{0.1015} \\[2pt]
fr3/s/static & 0.2826 & 0.1251 & -- & -- & 0.2709 & 0.1209 & \underline{0.2657} & 0.1163 & 0.2715 & \underline{0.1160} & \textbf{0.1532} & \textbf{0.0832} \\[2pt]
\hline
\end{tabular}
\vspace{1mm}
\begin{flushleft}
\hspace{3mm}\small \textbf{Note:} The best results of RMSE and S.D. are highlighted in bold, and the second best are underlined.
\end{flushleft}
\end{table*}

Our method, SR-SLAM, demonstrates strong competitiveness across a variety of challenging environments. In terms of ATE performance, our adaptive geometric constraint selection combined with the depth-assisted adaptive DBSCAN enables accurate removal of dynamic interference while effectively preserving usable features. In cases of insufficient features, direct method compensation enhances pose estimation robustness. This integrated strategy leads to state-of-the-art ATE results, achieving the best performance on both the most challenging motion sequence (fr3/w/half) and the low-dynamic sequence (fr3/s/static), and ranks first or second on three out of five sequences overall.

Regarding RPE, our reliability-based new keyframe selection and weighted optimization framework ensure consistency with various conditions. As a result, SR-SLAM outperforms all compared methods across nearly all sequences, often reducing errors by more than 50\% compared to the second-best approach. These improvements are particularly evident in sequences such as fr3/w/half and fr3/s/static, underscoring the effectiveness of our scene reliability-based optimization pipeline.

We also compare SR-SLAM with our baseline ORB-SLAM3 and three recent ORB-SLAM3 based dynamic methods: OVD-SLAM~\cite{OVD-SLAM}, O.D~\cite{O.D}, and HMC-SLAM~\cite{HMC-SLAM}, as shown in Table~\ref{tab:comprehensive_results based on orbslam3}. 
\begin{table*}[!htbp]
\centering
\caption{COMPARATIVE RESULTS OF DIFFERENT SLAM SYSTEMS BASED ON ORB-SLAM3.}
\label{tab:comprehensive_results based on orbslam3}
\small  
\vspace{-2mm}
\setlength{\tabcolsep}{8pt}
\renewcommand{\arraystretch}{1.3}
\begin{tabular}{l|c@{\hspace{8pt}}c|c@{\hspace{8pt}}c|c@{\hspace{8pt}}c|c@{\hspace{8pt}}c|c@{\hspace{8pt}}c|c@{\hspace{8pt}}c}
\hline
\rule{0pt}{3ex}
& \multicolumn{2}{c|}{ORB-SLAM3} & \multicolumn{2}{c|}{OVD-SLAM} & \multicolumn{2}{c|}{O.D} & \multicolumn{2}{c|}{HMC-SLAM} & \multicolumn{2}{c|}{OURS} & \multicolumn{2}{c}{Improvments} \\
\rule{0pt}{3ex} 
Sequences & RMSE & S.D. & RMSE & S.D. & RMSE & S.D. & RMSE & S.D. & RMSE & S.D. & RMSE & S.D. \\
\hline
\multicolumn{13}{c}{\textbf{ABSOLUTE TRAJECTORY ERROR (ATE)}} \\
\hline
\rule{0pt}{3ex} 
\hspace{-1mm}fr3/w/half & 0.0323 & 0.1141 & 0.0251 & 0.0122 & 0.0266 & 0.0159 & \textbf{0.0195} & \textbf{0.0099} & \underline{0.0242} & \underline{0.0120} & 25.08\% & 89.49\% \\[2pt]
fr3/w/xyz & 0.3781 & 0.2645 & 0.0153 & \underline{0.0074} & 0.0160 & \textbf{0.0062} & \underline{0.0152} & 0.0083 & \textbf{0.0150} & 0.0075 & 97.79\%  & 97.69\% \\[2pt]
fr3/w/rpy & 0.7642 & 0.4692 & 0.0531 & 0.0271 & 0.0295 & 0.0191 & \textbf{0.0257} & \textbf{0.0139} & \underline{0.0278} & \underline{0.0145} & 96.36\%  & 96.91\%  \\[2pt]
fr3/w/static & 0.3071 & 0.0722 & \underline{0.0071} & \underline{0.0028} & \textbf{0.0058} & \textbf{0.0028} & 0.0095 & 0.0059 & 0.0091 & 0.0037 & 97.04\% & 94.94\% \\[2pt]
fr3/s/static & 0.0109 & 0.0040 & 0.0081 & 0.0037 & 0.0062 & 0.0034 & \underline{0.0059} & \textbf{0.0026} & \textbf{0.0055} & \underline{0.0029} & 52.29\% & 47.27\% \\[2pt] 
\hline
\multicolumn{13}{c}{\textbf{METRIC TRANSLATIONAL DRIFT (T.RPE)}} \\
\hline
\rule{0pt}{3ex} 
\hspace{-1mm}fr3/w/half & 0.2298 & 0.1857 & 0.0275 & 0.0135 & \underline{0.0146} & \underline{0.0087} & 0.0217 & 0.0119 & \textbf{0.0129} & \textbf{0.0073} & 94.40\% & 96.07\% \\[2pt]
fr3/w/xyz & 0.2091 & 0.1722 & 0.0207 & 0.0099 & \underline{0.0175} & 0.0101 & 0.0189 & \underline{0.0089} & \textbf{0.0131} & \textbf{0.0078} & 93.75\% & 95.47\% \\[2pt]
fr3/w/rpy & 0.8161 & 0.3024 & 0.0414 & 0.0251 & \underline{0.0344} & \underline{0.0227} & 0.0432 & 0.0246 & \textbf{0.0194} & \textbf{0.0119} & 97.62\% & 96.06\% \\[2pt]
fr3/w/static & 0.1158 & 0.1002 & 0.0088 & 0.0039 & 0.0089 & 0.0055 & \underline{0.0082} & \underline{0.0035} & \textbf{0.0074} & \textbf{0.0043} & 93.61\% & 95.69\% \\[2pt]
fr3/s/static & 0.0103 & 0.0049 & 0.0094 & 0.0049 & \underline{0.0064} & \underline{0.0033} & 0.0077 & 0.0036 & \textbf{0.0049} & \textbf{0.0025} & 52.43\% & 48.98\% \\[2pt]
\hline
\multicolumn{13}{c}{\textbf{METRIC ROTATIONAL DRIFT (ROT. RPE)}} \\
\hline
\rule{0pt}{3ex} 
\hspace{-1mm}fr3/w/half & 4.3501 & 3.5526 & 0.7906 & \underline{0.3928} & - & - & \underline{0.7429} & 0.4256 & \textbf{0.4061} & \textbf{0.2163} & 90.67\% & 93.91\%  \\[2pt]
fr3/w/xyz & 3.9489 & 3.2623 & 0.6268 & 0.3831 & - & - & \underline{0.6218} & \underline{0.3822}& \textbf{0.3939} & \textbf{0.2752} & 90.02\% & 91.57\% \\[2pt]
fr3/w/rpy & 5.7739 & 3.9271 & 0.9266 & 0.5761 & - & - & \underline{0.8536} & \underline{0.4557 } & \textbf{0.6151} & \textbf{0.4693} & 89.37\% & 88.04\% \\[2pt]
fr3/w/static & 2.0924 & 1.7789 & 0.2437 & 0.1061 & - & - & \underline{0.2385} & \underline{0.1055} & \textbf{0.1900} & \textbf{0.1015} & 90.92\% & 94.29\% \\[2pt]
fr3/s/static & 0.3072 & 0.1299 & 0.2735 & 0.1183 & - & - & \underline{0.2605} & \underline{0.1129} & \textbf{0.1532} & \textbf{0.0832} & 50.13\% & 35.94\% \\[2pt]
\hline
\end{tabular}
\vspace{1mm}
\begin{flushleft}
\hspace{2mm}\small \textbf{Note:} The best results of RMSE and S.D. are highlighted in bold, and the second best are underlined.
\end{flushleft}
\end{table*}

In ATE, compared to the strong baseline, SR-SLAM achieves substantial improvements on both dynamic and static sequences, reducing ATE by over 95\% in highly dynamic cases (fr3/w/xyz, fr3/w/rpy) and up to 50\% in the static fr3/s/static. While other ORB-SLAM3 based dynamic methods also demonstrate improvements over the baseline, SR-SLAM consistently achieves either the best or second-best performance on four out of five sequences, securing the best ATE results in fr3/w/xyz (0.0150) and fr3/s/static (0.0055). These substantial gains result from our adaptive dynamic culling strategy that flexibly selects geometric constraints and employs depth-assisted adaptive DBSCAN to preserve usable features while removing dynamic interference. Additionally, our pose refinement integrates direct methods when feature-based estimation shows insufficient reliability, ensuring robust tracking even in diverse environments.

In RPE, our method achieves the best RPE performance across all sequences, with T.RPE errors reduced by over 90\% compared to ORB-SLAM3 and maintaining substantial advantages over competing dynamic methods. Compared to other ORB-SLAM3 based approaches, SR-SLAM consistently secures the best RPE results across all sequences, with errors often reduced by approximately half compared to the second-best method, for instance, achieving 0.0194 in fr3/w/rpy compared to O.D's 0.0344, 0.4061 in fr3/w/half versus HMC-SLAM's 0.7429. This systematic superiority stems from our weighted optimization combined with new keyframe selection, which enables frame-level optimization weighting based on actual scene conditions, ensuring superior relative pose accuracy across diverse environments.

\begin{figure*}[!htbp]
  \begin{center}
  \includegraphics[width=\linewidth]{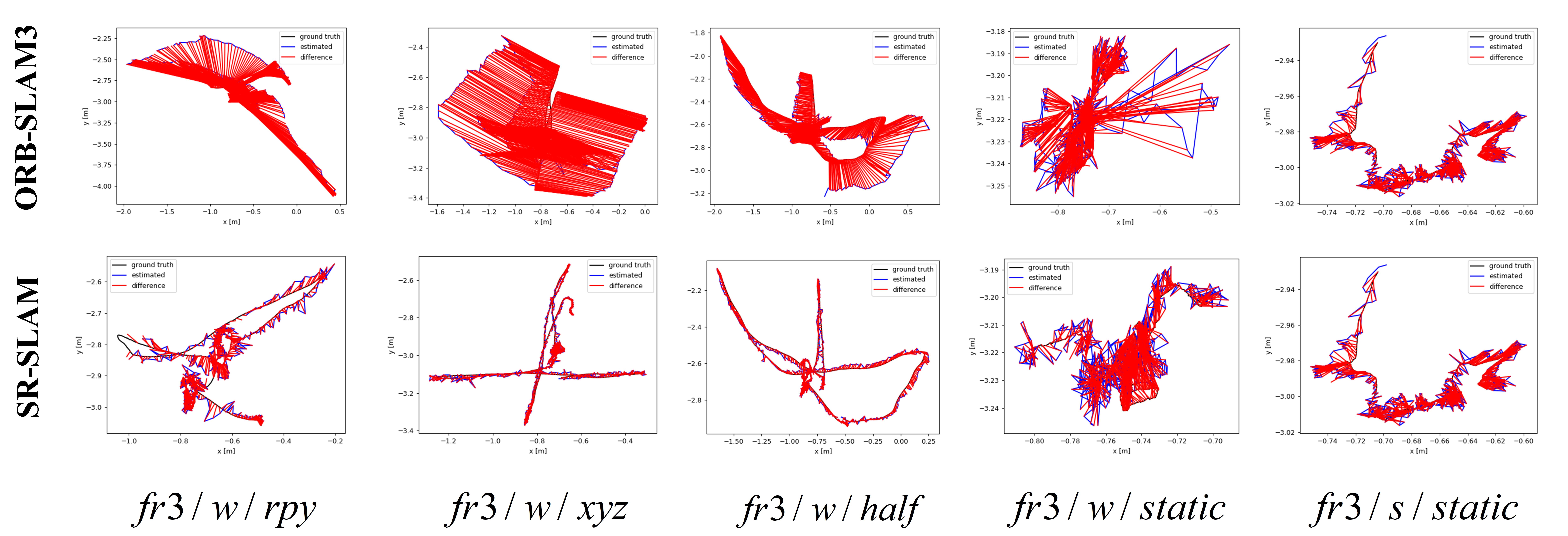}
  \caption{ATE results of SR-SLAM and ORB-SLAM3 running in five sequences, where the black line represents the true trajectory, the blue line represents the trajectory estimated by the algorithm, and the red line represents the difference between the estimated and true values}
  \label{fig:Trajectory_compare}
  \end{center}
\end{figure*}

\begin{figure*}[!htbp]
  \begin{center}
  \includegraphics[width=\linewidth]{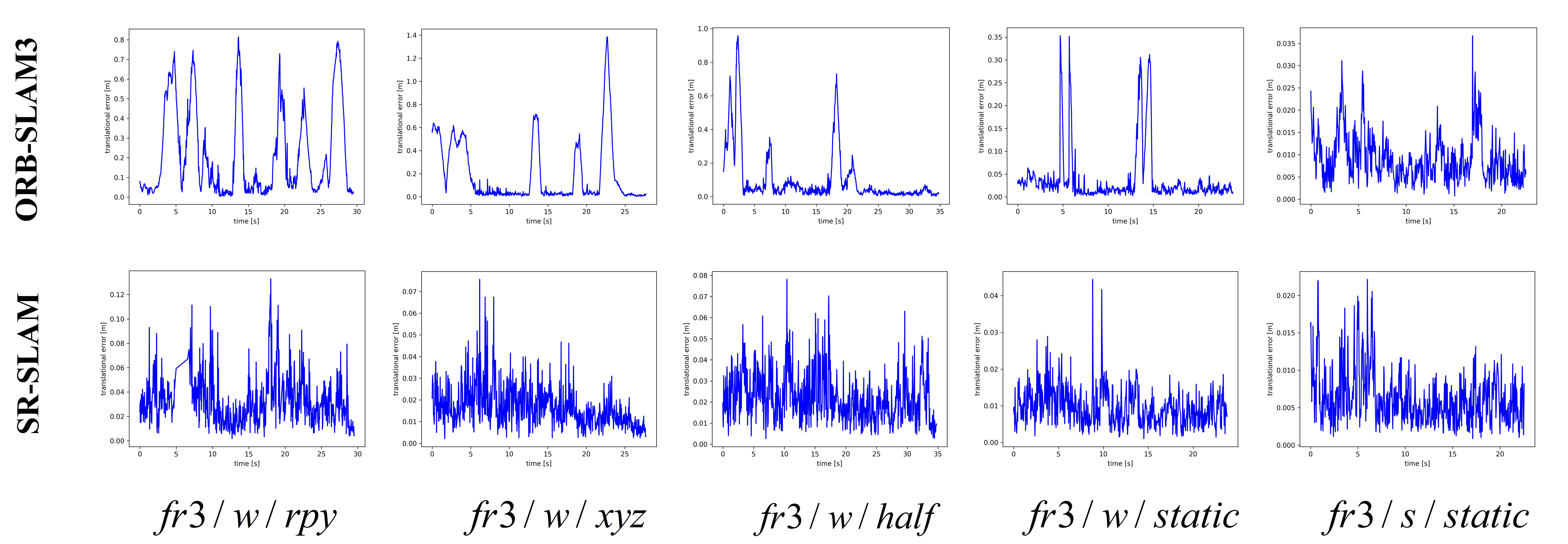}
  \caption{RPE results of SR-SLAM and ORB-SLAM3 running in five sequences, where the blue lines represent the RPE results for each time point.}
  \label{fig:Traj_ape}
  \end{center}
\end{figure*}

Fig.~\ref{fig:Trajectory_compare} and Fig.~\ref{fig:Traj_ape} show the estimated trajectories and RPE error between ORB-SLAM3 and our proposed SR-SLAM across five different sequences. It is evident that ORB-SLAM3 suffers from complete tracking failures in the first four dynamic sequences, producing chaotic trajectories with errors exceeding 0.35–1.4 meters. In contrast, SR-SLAM maintains stable and accurate tracking, with errors consistently kept below approximately 0.1 meters. Furthermore, in some challenging dynamic sequences such as fr3/w/xyz and fr3/w/half, the estimated trajectories of SR-SLAM are almost perfectly aligned with the ground truth, demonstrating the robustness of our reliability-based approach in diverse environments.

\subsection{Evaluation on the Bonn RGB-D Dataset}
The BONN dataset, provided by the University of Bonn in 2019 and captured using an ASUS Xtion Pro LIVE camera, contains scenes that are more complex than those in the TUM dataset. Therefore, we further validated our method on the BONN dataset. In detail, the crowd sequence presents three individuals walking randomly indoors. The moving\_no\_box sequence demonstrates a person transferring a box from the floor to a table and back. The person\_tracking sequence focuses on the camera tracking a single walking person, while the synchronous sequence shows two people moving in precisely polar directions at the same time.
\begin{table*}[!htbp]
\centering
\caption{RESULTS OF ATE ON BONN DATASET.}
\label{tab:ATE_BONN}
\small  
\vspace{-2mm}
\setlength{\tabcolsep}{8pt}
\renewcommand{\arraystretch}{1.3}
\begin{tabular}{l|c@{\hspace{8pt}}c|c@{\hspace{8pt}}c|c@{\hspace{8pt}}c|c@{\hspace{8pt}}c|c@{\hspace{8pt}}c|c@{\hspace{8pt}}c}
\hline
\rule{0pt}{3ex}
& \multicolumn{2}{c|}{ORB-SLAM3} & \multicolumn{2}{c|}{YOLO-SLAM} & \multicolumn{2}{c|}{SG-SLAM} & \multicolumn{2}{c|}{SSF-SLAM} & \multicolumn{2}{c|}{OURS} & \multicolumn{2}{c}{Improvments} \\
\rule{0pt}{3ex} 
\hspace{-1mm}Sequences & RMSE & S.D. & RMSE & S.D. & RMSE & S.D. & RMSE & S.D. & RMSE & S.D. & RMSE & S.D. \\
\hline
\rule{0pt}{3ex} 
\hspace{-1mm}crowd & 0.4960 & 0.4130 & 0.0330 & - & \underline{0.0234} & 0.0143 & \textbf{0.0228} & \textbf{0.0117} & 0.0240 & \underline{0.0137} & 95.16\% & 96.68\%\\[2pt]
crowd2 & 0.9890 & 0.5970 & 0.4230 & - & 0.0584 & 0.0406 & \textbf{0.0339} & \textbf{0.0196} & \underline{0.0551} & \underline{0.0387} & 94.43\% & 96.68\% \\[2pt]
crowd3 & 0.5050 & 0.3280 & 0.0690 & - & \underline{0.0319} & 0.0219 & 0.0363 & \underline{0.0213} & \textbf{0.0310} & \textbf{0.0178} & 94.57\% & 96.68\% \\[2pt]
moving\_no\_box & 0.3200 & 0.0900 & 0.0270 & - & \textbf{0.0192} & \textbf{0.0081} & \underline{0.0212} & \underline{0.0099} & 0.0346 & 0.0260 & 89.19\% & 71.11\% \\[2pt]
moving\_no\_box2 & 0.0390 & 0.0150 & 0.0350 & - & \textbf{0.0299} & \underline{0.0119} & \underline{0.0321} & 0.0135 & 0.0332 & \textbf{0.0106} & 14.87\% & 29.33\% \\[2pt] 
person\_tracking & 0.6920 & 0.3390 & 0.1570 & - & \underline{0.0400} & \underline{0.0139} & \textbf{0.0370} & \textbf{0.0136} & 0.0583 & 0.0186 & 91.58\% & 94.51\%\\[2pt] 
person\_tracking2 & 0.7800 & 0.4910 & 0.0370 & - & \underline{0.0376} & \underline{0.0154} & \textbf{0.0369} & \textbf{0.0136} & 0.0852 & 0.0449 & 89.08\% & 90.86\%\\[2pt] 
synchronous & 0.8090 & 0.4400 & \textbf{0.0140} & - & 0.3229 & \underline{0.1824} & 0.0280 & 0.0232 & \underline{0.0153} & \textbf{0.0116} & 98.11\% & 97.36\%\\[2pt] 
synchronous2 & 1.2970 & 0.1580 & \textbf{0.0070} & - & 0.0164 & 0.0126 & 0.0275 & \underline{0.0174} & \underline{0.0110} & \textbf{0.0056} & 99.15\% & 96.46\%\\[2pt] 
\hline
\rule{0pt}{3ex} 
\hspace{-1mm}Average & 0.5597 & 0.3190 & 0.0891 & - & 0.0644 & 0.0357 & \textbf{0.0306} & \textbf{0.0159} & \underline{0.03863} & \underline{0.02083} & - & - \\[2pt] 
\hline
\end{tabular}
\vspace{1mm}
\begin{flushleft}
\hspace{2mm}\small \textbf{Note:} The best results of RMSE and S.D. are highlighted in bold, and the second best are underlined.
\end{flushleft}
\end{table*}

The evaluation and comparison results of nine dynamic scene sequences are shown in Table~\ref{tab:ATE_BONN}. Our method achieves significant improvements over the baseline ORB-SLAM3, with error reductions of around 90\% or higher in most sequences except for moving\_no\_box2, where the improvement is relatively limited. Compared with the other three methods, our approach ranks either first or second in 10 out of 20 indicators (RMSE and S.D. combined), accounting for half of the total metrics. The relatively lower performance on the person\_tracking and moving\_no\_box sequences is mainly due to the characteristics of our dynamic filtering and pose completion strategy. 

In the person\_tracking case, where the camera undergoes continuous long-range motion while dynamic content dominates the scene, our adaptive dynamic culling strategy deliberately applies maximum removal of all potential dynamic regions to minimize contamination. This aggressive filtering leads to severely limited static information, causing most frames to be marked as unreliable and consistently triggering our direct pose refinement module. However, due to the sustained camera movement, the historical reference information becomes less reliable, and the photometric and depth consistency assumptions underlying direct method optimization are frequently violated between reference frames and current observations. Consequently, the continuous reliance on direct method compensation under these challenging conditions results in degraded performance compared to scenarios with more stable camera configurations or sufficient static features. For moving\_no\_box sequences, the slight performance drop occurs because our system relies primarily on geometric constraints to identify undefined dynamic objects (such as the moving box). In low-motion scenarios, our adaptive dynamic culling strategy employs the depth-assisted DBSCAN method to preserve more usable features and avoid over-removal. However, this conservative approach can erroneously retain features on undefined dynamic objects. Consequently, these incorrectly preserved dynamic features introduce residual motion inconsistencies that affect pose estimation accuracy. Nevertheless, our method excels in complex multi-dynamic scenarios such as crowd and synchronous sequences, which involve intricate dynamic interactions. Most importantly, while our ATE is not always optimal, our system consistently avoids the catastrophic failures observed in YOLO-SLAM and SG-SLAM on challenging sequences like person\_tracking and crowd2. This demonstrates that our reliability-based approach maintains stable and robust performance across diverse dynamic environments, prioritizing system reliability over peak performance in extreme scenarios.

\subsection{Ablation Experiment}
Our method, SR-SLAM, leverages the advantages of a scene reliability assessment mechanism, adaptive dynamic culling strategies, pose refinement, new keyframe selection, and optimization mechanisms to achieve intelligent dynamic feature selection and frame-level adjustment across diverse scenarios. To validate the effectiveness of these individual modules, we designed comprehensive ablation experiments. SR-SLAM (SP) represents our baseline ORB-SLAM3, incorporating the scene reliability assessment mechanism and pose refinement strategy. SR-SLAM (SP+D) extends SR-SLAM (SP) by integrating our adaptive dynamic culling strategies. SR-SLAM (SP+B) builds upon SR-SLAM (SP) with the addition of our new keyframe selection and optimization mechanism for the backend. Finally, SR-SLAM represents the complete framework incorporating all proposed components. The ATE experimental results on the TUM dataset are presented in Table~\ref{tab:Ablation Experiments_ATE}.

The experimental results demonstrate that, compared to our baseline ORB-SLAM3, all tested sequences achieve remarkable improvements across both dynamic and static scenarios. On average, SR-SLAM (SP), SR-SLAM (SP+D), SR-SLAM (SP+B), and SR-SLAM achieve improvements of 92.89\%, 93.49\%, 93.57\%, and 93.89\% in RMSE, respectively, and 94.05\%, 94.71\%, 94.83\%, and 95.09\% in S.D. compared to the baseline. Furthermore, our complete SR-SLAM framework achieves the best or second-best performance in 10 out of 12 evaluation metrics when compared to the ablated variants, and obtains the best performance in average metrics. This demonstrates that SR-SLAM maintains high precision while exhibiting excellent adaptability and balanced performance across diverse scenarios.
\begin{table*}[!htbp]
\centering
\caption{Ablation Experiments}
\label{tab:Ablation Experiments_ATE}
\small
\vspace{-2mm}
\setlength{\tabcolsep}{10pt}
\begin{tabular}{l|cc|cc|cc|cc|cc}
\hline
\rule{0pt}{2ex} 
& \multicolumn{10}{c}{\textbf{ABSOLUTE TRAJECTORY ERROR (ATE)}} \\
\cline{2-11}
\rule{0pt}{2ex}
\hspace{-1mm}Sequences & \multicolumn{2}{c|}{ORB-SLAM3} & \multicolumn{2}{c|}{SRR-SLAM(SP)} & \multicolumn{2}{c|}{SRR-SLAM(SP+D)} & \multicolumn{2}{c|}{SRR-SLAM(SP+B)} & \multicolumn{2}{c}{SRR-SLAM}\\
& RMSE & S.D. & RMSE & S.D. & RMSE & S.D. & RMSE & S.D. & RMSE & S.D.  \\
\hline
\rule{0pt}{2ex}
\hspace{-1mm}fr3/w/half & 0.0323 & 0.1141 & 0.0286 & 0.0150  & 0.0276 & 0.0134 & \underline{0.0253} & \underline{0.0128} & \textbf{0.0242} & \textbf{0.0120}  \\
fr3/w/xyz & 0.3781 & 0.2645 & 0.0152 & 0.0077 & 0.0153 & 0.0082 & \textbf{0.0148} & \textbf{0.0070} & \underline{0.0150} & \underline{0.0075}  \\
fr3/w/rpy & 0.7642 & 0.4692 & 0.0302 & 0.0164 & \underline{0.0285} & \underline{0.0155} & 0.0288 & 0.0160 & \textbf{0.0278} & \textbf{0.0145}   \\
fr3/w/static & 0.3071 & 0.0722 & \underline{0.0087} & \underline{0.0038} & \textbf{0.0084} & 0.0040 & 0.0093 & 0.0041 & 0.0091 & \textbf{0.0037} \\
fr3/s/static & 0.0109 & 0.0040 & 0.0067 & \underline{0.0033} & \underline{0.0063} & \textbf{0.0029} & \textbf{0.0059} & \textbf{0.0029} & \textbf{0.0055} & 0.0029 \\
fr3/s/xyz & 0.0093 & 0.0048 & 0.0172 & 0.0091 & \underline{0.0115} & \underline{0.0052} & 0.0125 & 0.0053 & \textbf{0.0102} & \textbf{0.0048} \\
\hline
\rule{0pt}{2ex}
\hspace{-1.5mm}Average & 0.2503 & 0.1548 & 0.0178 & 0.0092 & 0.0163 & 0.0082 & \underline{0.0161} & \underline{0.0080} & \textbf{0.0153} & \textbf{0.0076} \\
Improvments(\%) & - & - & 92.89 & 94.05 & 93.49 & 94.71 & \underline{93.57} & \underline{94.83} & \textbf{93.89} & \textbf{95.09} \\
\hline
\end{tabular}
\vspace{0.2mm}
\begin{flushleft}
\hspace{1mm}\scriptsize \textbf{Note:} The best results of RMSE and S.D. are highlighted in bold.
\end{flushleft}
\end{table*}
\begin{figure*}[!htbp]
  \begin{center}
  \includegraphics[width=\linewidth]{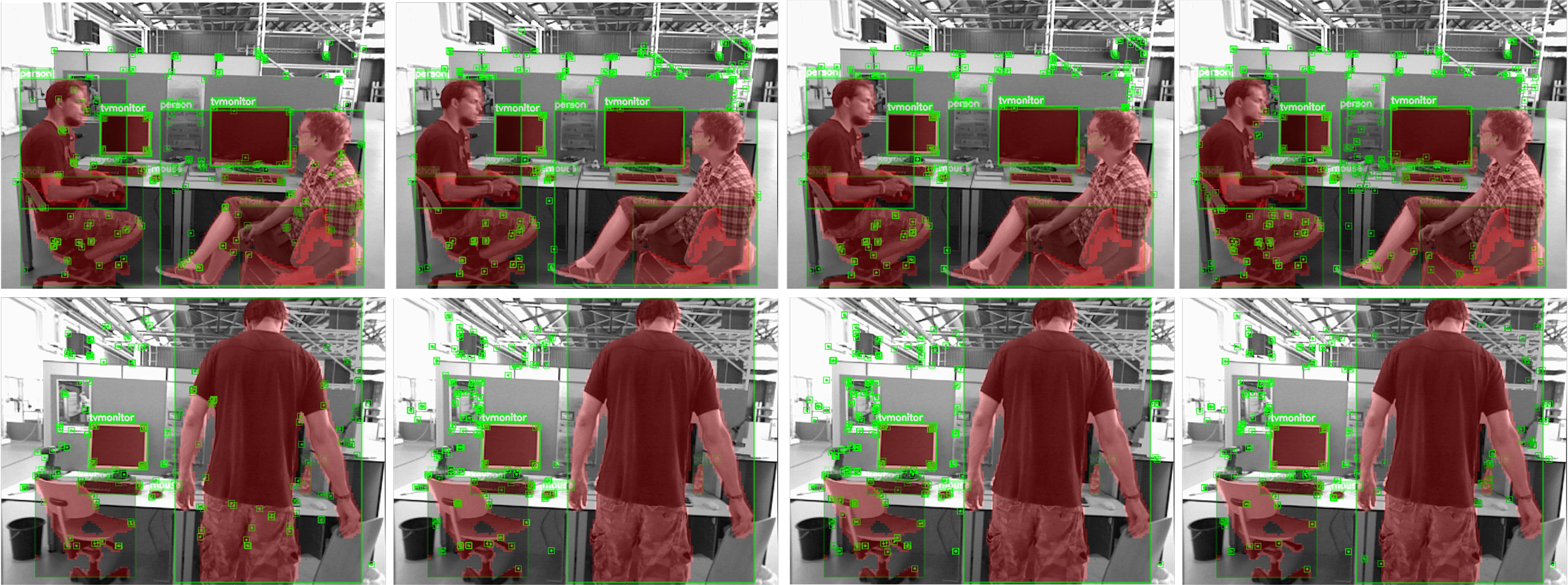}
  \caption{Ablation study results for dynamic feature culling. The first row shows results in a low dynamic scene, while the second row corresponds to a moderately dynamic environment. From left to right: culling effects using (a) only geometric constraints, (b) only detection, (c) detection with geometric constraints and (d) our depth-assisted adaptive DBSCAN.}
  \label{fig:removal ablation}
  \end{center}
\end{figure*}
Additionally, we conducted ablation studies on our adaptive dynamic culling strategies in low dynamic and moderately dynamic scenes, as illustrated in Fig~\ref{fig:removal ablation}. As shown in (a), relying solely on geometric constraints results in substantial residual motion points in both scenes, significantly impacting tracking accuracy. While approaches using detection only (b) and the hybrid method combining geometric constraints and detection (c) successfully remove dynamic points from moving objects in moderately dynamic scenes, they exhibit over-removal tendencies in low dynamic scene, erroneously eliminating many useful static feature points and consequently reducing the richness of available features. In contrast, our adaptive dynamic culling strategy (d) performs the scene-adaptation in potential dynamic region selection and final dynamic feature removal. As demonstrated in (d), our method achieves precise removal of dynamic object features while maximally preserving usable features in potential dynamic regions. To clarify, features on the obvious dynamic object (person) are effectively eliminated, while rich feature points are preserved around the person's contour region in both scenes. Moreover, our method successfully extracts usable features even in narrow regions between the person's arms and torso, highlighting the flexibility and precision of our removal strategy.

\subsection{Runtime Analysis}
Regarding computational efficiency, we conducted separate timing tests for each proposed module and compared the overall runtime with our baseline method ORB-SLAM3 on representative sequences from the TUM dataset. The results are presented in Table~\ref{tab:Each Proposed Module Runtime} and Table~\ref{tab:Overall Runtime Analysis}. 

As shown in Table~\ref{tab:Each Proposed Module Runtime}, the detection and geometry constraints module (De.G) represents the primary computational bottleneck, consuming 17.47ms per frame on average across all sequences. The scene assessment module (S) requires 8.57ms per frame, while our adaptive pose refinement (P) demonstrates selective activation ranging from 0ms in stable sequences to 3.70ms in challenging scenarios. Notably, our dynamic culling (D) and new keyframe selection and optimization mechanism (B) modules introduce minimal overhead below 0.01ms, highlighting the computational efficiency of our frame-level processing strategy.
\begin{table*}[!htbp]
\centering
\caption{Each Proposed Module Runtime}
\label{tab:Each Proposed Module Runtime}
\small
\vspace{-2mm}
\setlength{\tabcolsep}{10pt}
\begin{tabular}{l|c|c|c|c|c}
\hline
\rule{0pt}{2ex} 
& \multicolumn{4}{c}{\textbf{Each Proposed Module Runtime (frame/ms)}} \\
\cline{2-6}
\rule{0pt}{2ex}
\hspace{-1mm}Sequences & \multicolumn{1}{c|}{SRR-SLAM(De.G)} & \multicolumn{1}{c|}{SRR-SLAM(S)} & \multicolumn{1}{c|}{SRR-SLAM(P)} & \multicolumn{1}{c|}{SRR-SLAM(D)} & \multicolumn{1}{c}{SRR-SLAM(B)}\\  
\hline
\rule{0pt}{2ex}
\hspace{-1mm}fr3/w/half & 16.4245 & 8.9248 & 2.9576 & 0.0022 & $1.9782\text{e}^{-6}$
 \\
fr3/w/xyz & 17.0767 & 8.7871 & 1.4511 & 0.0025 & $2.5188\text{e}^{-6}$  \\
fr3/w/rpy & 16.4930 & 8.8085 & 3.5704 & 0.0013 & $2.3148\text{e}^{-6}$  \\
fr3/w/static & 18.9052 & 8.3957 & 0.0000 & 0.0071 & 0.0000\\
fr3/s/static & 17.7381 & 7.8601 & 0.0000 & 0.0129 & 0.0000 \\
fr3/s/xyz & 18.1923 & 8.6701 & 2.5213 & 0.0070 & 0.0000  \\
\hline
\rule{0pt}{2ex}
\hspace{-1.5mm}Average & 17.4716 & 8.5744 & 2.0834 & 0.0055 & $1.1353\text{e}^{-6}$  \\
\hline
\end{tabular}
\vspace{0.2mm}
\end{table*}

Following the module-level analysis, Table~\ref{tab:Overall Runtime Analysis} presents the overall system performance comparison. SR-SLAM maintains an average processing time of approximately 0.07 seconds per frame, achieving around 14 FPS, with consistent performance across different sequences (0.0704-0.0712s per frame). As identified in the module analysis, the detection and geometry constraints module constitutes the primary computational bottleneck, contributing most of the processing overhead compared to the baseline. Nevertheless, this trade-off between robust dynamic processing and computational efficiency is justified for applications where tracking precision is prioritized over maximum speed, making the system suitable for real-world applications requiring robust SLAM performance.
\begin{table}[!htbp]
\centering
\caption{Overall Runtime Analysis}
\label{tab:Overall Runtime Analysis}
\small
\vspace{-2mm}
\setlength{\tabcolsep}{10pt}
\begin{tabular}{l|c|c}
\hline
\rule{0pt}{2ex}
& \multicolumn{2}{c}{\textbf{Overall Runtime (frame/s)}} \\
\cline{2-3}
\rule{0pt}{2ex}
\hspace{-1mm}Sequences & \multicolumn{1}{c|}{ORB-SLAM3} & \multicolumn{1}{c}{SRR-SLAM} \\
\hline
\rule{0pt}{2ex}
\hspace{-1mm}fr3/w/half & 0.0371 & 0.0712    \\
fr3/w/xyz & 0.0359 & 0.0707   \\
fr3/w/rpy & 0.0369 & 0.0706    \\
fr3/w/static & 0.0364 & 0.0704  \\
fr3/s/static & 0.0358 & 0.0702  \\
fr3/s/xyz & 0.0359 & 0.0704   \\
\hline
\rule{0pt}{2ex}
\hspace{-1.5mm}Average & 0.0363 & 0.0705  \\
\hline
\end{tabular}
\vspace{0.2mm}
\end{table}
\subsection{Experiments in Real Scenarios}
\begin{figure}[!htbp]
  \begin{center}
  \includegraphics[width=\linewidth]{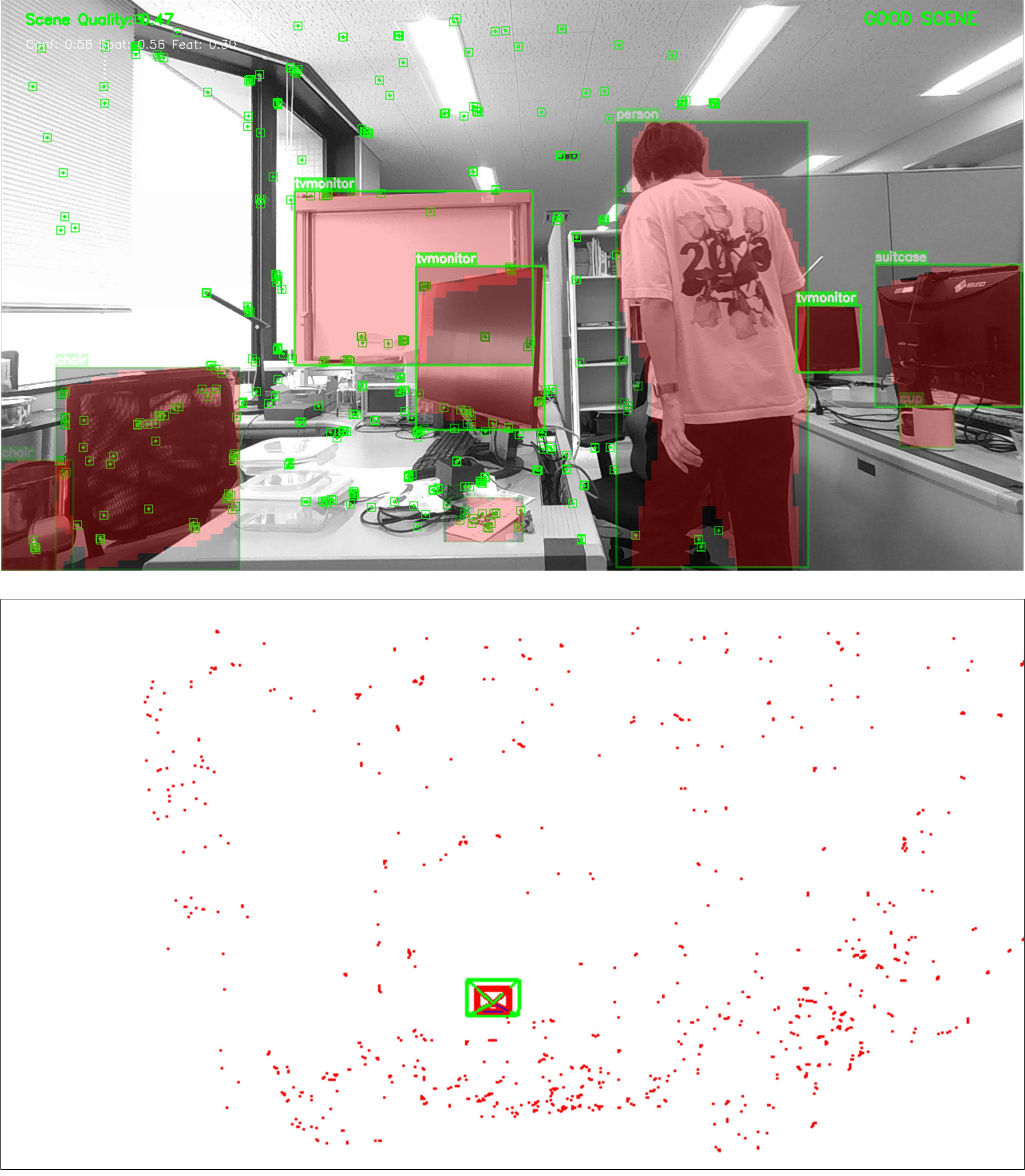}
  \caption{Performance of SR-SLAM in a real environment. The upper panel shows the current frame with our scene reliability assessment in action, displaying detected dynamic objects and reliability metrics. The lower panel presents the sparse mapping results and camera trajectory estimation.}
  \label{fig:real_performance}
  \end{center}
\end{figure}

In real-world scenarios, we first tested SR-SLAM using a fixed Intel RealSense D455 camera with a single moving person in an indoor environment. As shown in Fig.~\ref{fig:real_performance}, the upper panel displays SR-SLAM's real-time performance with detected dynamic objects (green boxes), segmented masks (red), scene result and reliability (0.47) in green, and individual metrics ($R_{conf}$, $R_{feature}$...) in white. The lower panel shows mapping results and camera pose. Our method demonstrates robust performance with effective dynamic feature removal while preserving usable features around moving objects, even in narrow spaces between a person's legs. The camera pose estimation remains highly stable for the fixed camera setup, and the trajectory comparisons presented in Fig.~\ref{fig:real_traj}.(f).   

\begin{figure}[!htbp]
  \begin{center}
  \includegraphics[width=\linewidth]{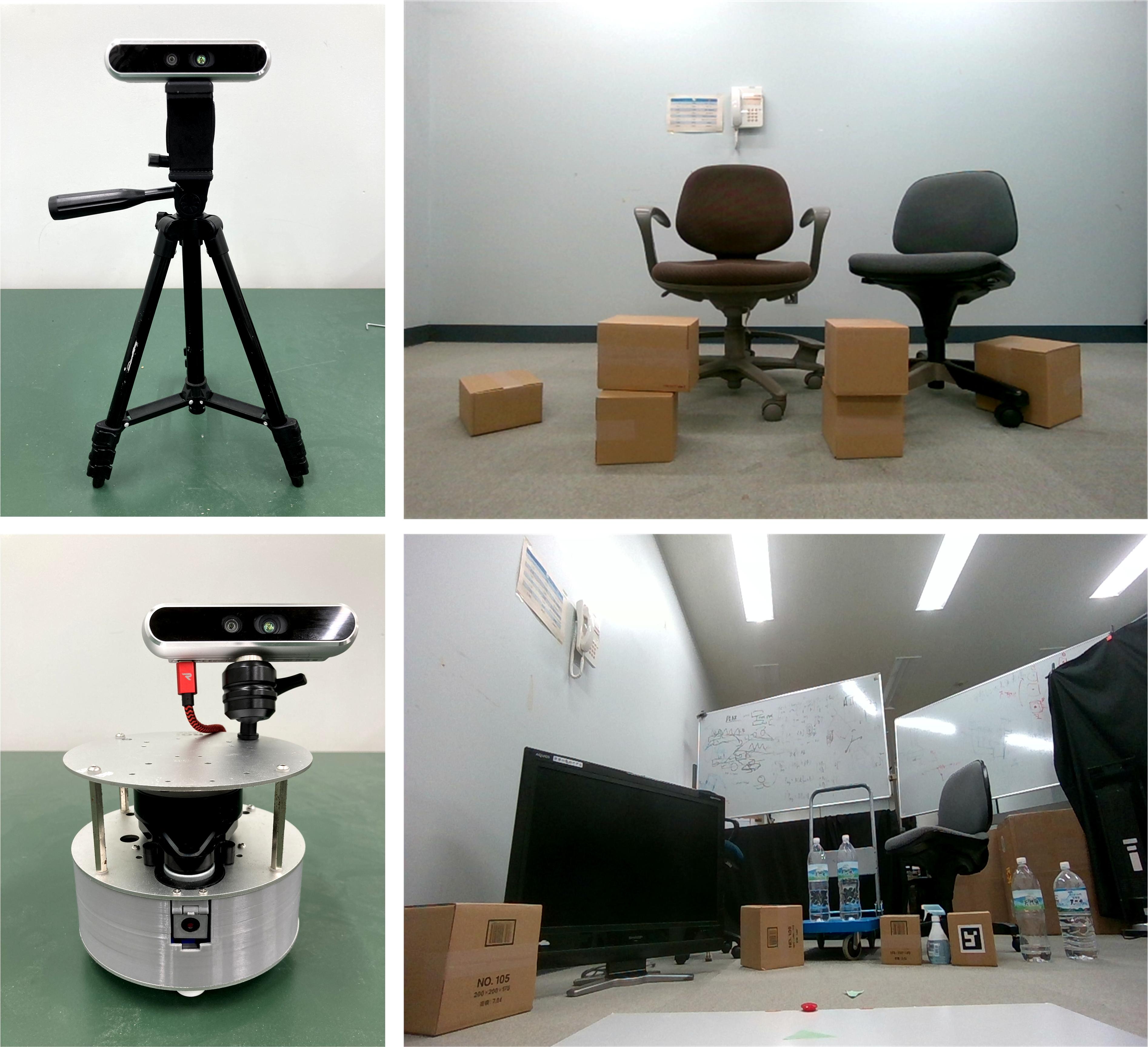}
  \caption{Real experimental settings. First row: images 1-2 show a fixed camera setup and environmental configuration, respectively. Second row: images 1-2 show the mobile robot setup and environmental configuration respectively.}
  \label{fig:real setting}
  \end{center}
\end{figure}

\begin{figure*}[!htbp]
  \begin{center}
  \includegraphics[width=\linewidth]{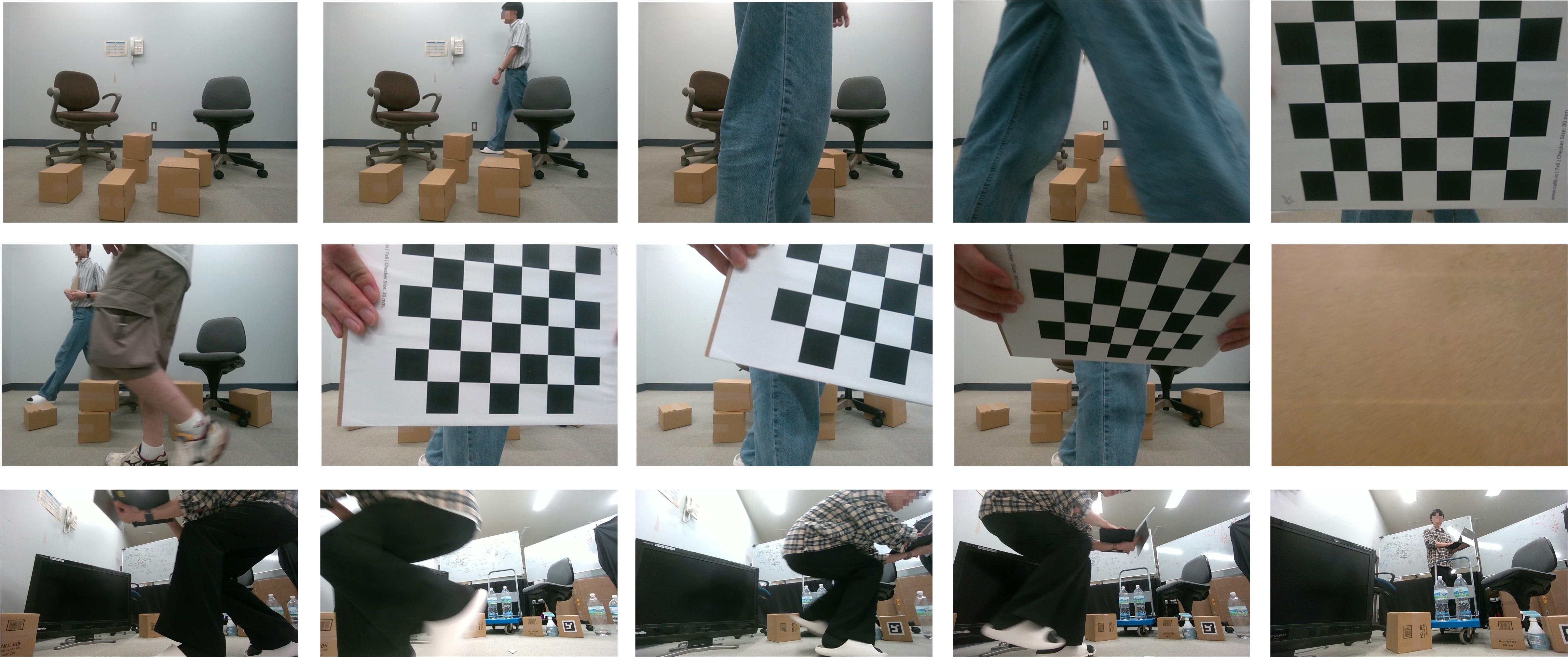}
  \caption{Environment settings. In fixed camera setting, four experiments: (a) first row, first image shows static environment; (b) first row, images 2-3 contain dynamic objects and minor occlusions; (c) first row, images 4-6 contain dynamic objects and major occlusions; (d) all images in second row contain multiple dynamic objects, multi-angle major occlusions and complete occlusions. In the mobile robot setting, experiment (e): all images in the third row, from left to right, include dynamic major occlusions, near dynamic objects, and distant dynamic objects.}
  \label{fig:real_factor_setting}
  \end{center}
\end{figure*}

\begin{figure}[!htbp]
  \begin{center}
  \includegraphics[width=\linewidth]{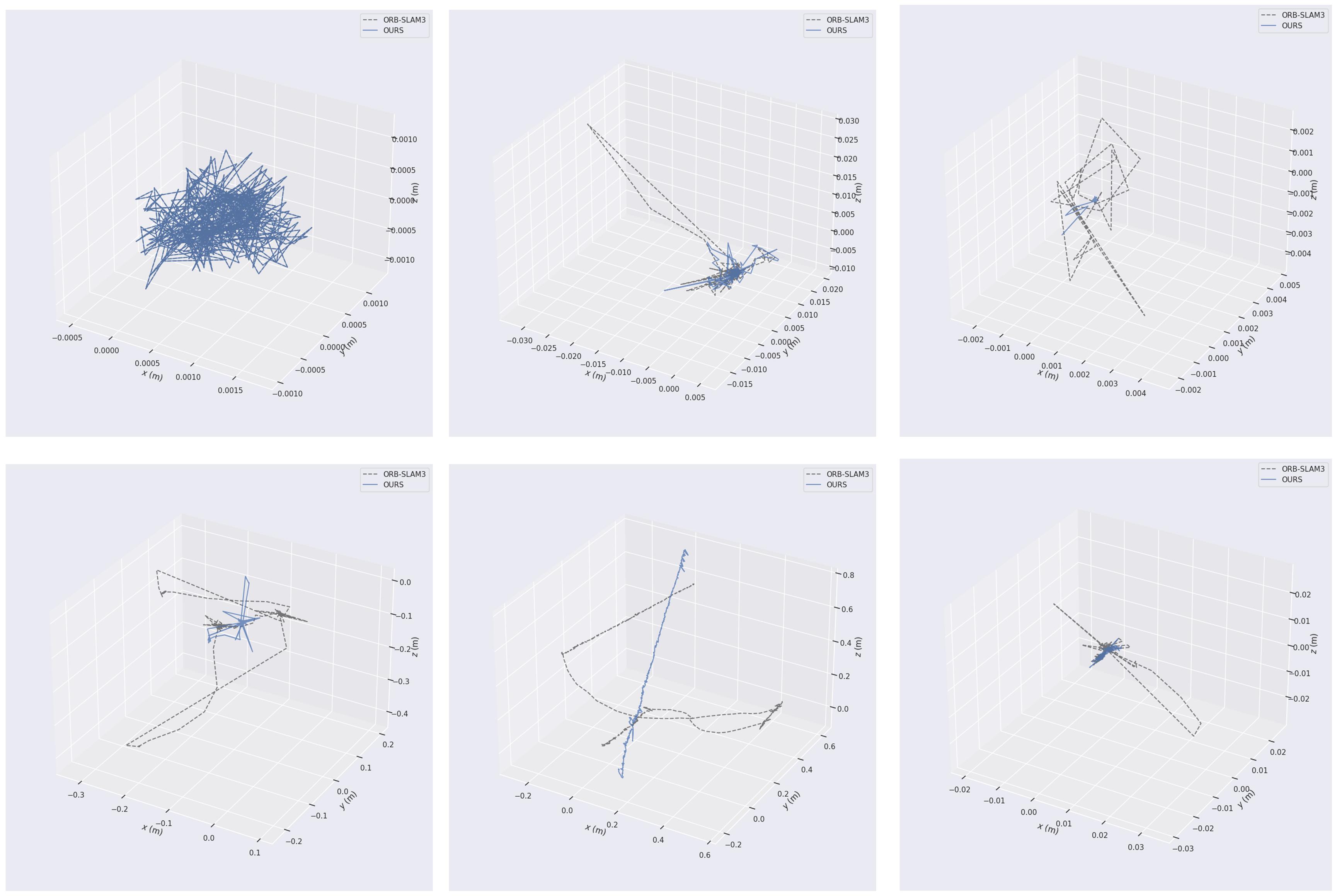}
  \caption{Comparison of camera trajectories in real-world experiments. Blue trajectories represent SR-SLAM results, while gray trajectories represent ORB-SLAM3 results. The first three images in the first row correspond to scenarios (a), (b), and (c) in Fig.~\ref{fig:real setting}, respectively. The first and second images in the second row represent scenarios (d) and (e) in Fig.~\ref{fig:real setting}, respectively. The last image represents the test results from Fig.~\ref{fig:real_performance}.}
  \label{fig:real_traj}
  \end{center}
\end{figure}

To further validate SR-SLAM, we constructed experimental setups for both fixed camera and mobile robot (equipped with Intel RealSense D455) configurations, as shown in Fig.~\ref{fig:real setting}. In terms of environmental design, as illustrated in Fig.~\ref{fig:real_factor_setting}, we progressively introduced complexity factors starting from a static environment with a fixed camera in (a). In (b), we added a single dynamic object with minor occlusions. In (c), major dynamic and static occlusions were introduced 
to (b). In (d), we incorporated multiple dynamic objects, partial occlusions from different positions and angles, and complete occlusions. For the mobile robot scenario in (e), as the robot moved in a straight line, we created sustained dynamic large-scale occlusions with a person moving from left to right, along with close-range dynamic objects and distant dynamic object disturbances.

Under these real-world conditions, we conducted camera trajectory comparisons between SR-SLAM and baseline ORB-SLAM3, as shown in Fig.~\ref{fig:real_traj}. The gray trajectories represent ORB-SLAM3 results, while the blue trajectories represent SR-SLAM results. Subfigures (a), (b), (c), (d), and (e) correspond to the fixed camera and mobile robot scenarios defined in Fig.~\ref{fig:real setting}, while (f) represents the test scenario from Fig.~\ref{fig:real_performance}.

In the static environment (a), SR-SLAM's trajectory is completely consistent with ORB-SLAM3, demonstrating that our scene reliability assessment correctly identifies stable conditions and maintains baseline performance without unnecessary processing overhead. However, in complex dynamic environments (b), (c), (d), and (f), ORB-SLAM3's trajectories exhibit significant divergence and tracking failures, while SR-SLAM maintains stable trajectories near (0,0,0). Most notably, in the mobile robot straight-line motion scenario (e), ORB-SLAM3 completely diverges and fails to capture the camera's movement trend, whereas SR-SLAM maintains a straight-line trajectory even under simultaneous camera motion and complex dynamic environments. These superior results arise from our adaptive processing framework, which handles dynamic interference, maintains robust pose estimation under occlusion, and adjusts observation contributions based on scene conditions. Moreover, reliability-based keyframe selection ensures that only stable frames are retained for optimization.

These diverse real-world experiments demonstrate that our unified scene reliability assessment framework enables SR-SLAM to maintain robust adaptability across diverse environments while preserving high precision, further validating that the algorithm is highly effective in diverse environments.

\section{Conclusion}
We proposed SR-SLAM, a scene-reliability based RGB-D SLAM framework designed to overcome the limitations of existing feature-based methods in diverse environments. By integrating object detection confidence, spatial distribution, feature quality, depth quality, and historical references, we established a unified and adaptive scene reliability assessment mechanism at the frame level. Based on this assessment, SR-SLAM dynamically adjusts key components of the SLAM pipeline, including dynamic region selection, feature removal, pose refinement, keyframe selection, and optimization according to scene conditions.  

Extensive evaluations on public datasets (TUM and BONN) demonstrate that SR-SLAM consistently outperforms state-of-the-art dynamic SLAM systems, achieving significant improvements in both ATE and RPE metrics, particularly in complex settings. Furthermore, experiments in a real-world environment with various settings show that our method can be applied to diverse environments.


Future work will focus on incorporating additional environmental factors such as illumination variations~\cite{Illumination-Adaptive, QualiSLAM}, integrating more efficient detection modules for improved real-time performance~\cite{SG-SLAM, SSF-SLAM}, and introducing learning-based parameter adaptation for enhanced environmental adaptivity.


 





\end{document}